\documentclass{article} 
\usepackage{iclr2025_conference,times}

\usepackage{amsmath,amsfonts,bm}

\def\eqref#1{equation~\ref{#1}}

\def\1{\bm{1}}

\DeclareMathAlphabet{\mathsfit}{\encodingdefault}{\sfdefault}{m}{sl}
\SetMathAlphabet{\mathsfit}{bold}{\encodingdefault}{\sfdefault}{bx}{n}

\usepackage{hyperref}
\usepackage{xcolor}
\usepackage{url}
\usepackage{bm}
\usepackage{wrapfig}
\usepackage{graphicx}
\usepackage{booktabs}
\usepackage{tablefootnote}
\usepackage{threeparttable}
\usepackage{graphics}
\usepackage{multirow}
\usepackage{multicol}
\usepackage{subcaption}
\usepackage{cleveref}

\usepackage{xspace}

\usepackage{pgfplots}
\usepackage{pgfplotstable}
\pgfplotsset{compat=1.17}
\usetikzlibrary{patterns}

\title{RecTable: Fast Modeling Tabular Data with Rectified Flow}

\author{Masane Fuchi \& Tomohiro Takagi \\
Meiji University\\
\texttt{ce235031@meiji.ac.jp} 
}

\iclrfinalcopy 
\begin{document}

\maketitle

\begin{abstract}
Score-based or diffusion models generate high-quality tabular data, surpassing GAN-based and VAE-based models. However, these methods require substantial training time. In this paper, we introduce RecTable, which uses the rectified flow modeling, applied in such as text-to-image generation and text-to-video generation. RecTable features a simple architecture consisting of a few stacked gated linear unit blocks. Additionally, our training strategies are also simple, incorporating a mixed-type noise distribution and a logit-normal timestep distribution. Our experiments demonstrate that RecTable achieves competitive performance compared to the several state-of-the-art diffusion and score-based models while reducing the required training time. Our code is available at \url{https://github.com/fmp453/rectable}.
\end{abstract}

\begin{wrapfigure}[23]{r}{0.48\textwidth}
   \centering
   \resizebox{0.48\textwidth}{!}{
   \begin{tikzpicture}
      \begin{axis}[
           xlabel={Training Time (sec)},
           ylabel={AUC},
           xmin=1000, xmax=30000,
           xmode=log,
           ymin=0.77, ymax=0.95,
           legend pos=south west,
           grid=major
      ]

      \draw [pattern=north east lines, pattern color=blue!20] (axis cs:1000,0.90) rectangle (axis cs:3162,0.95);
      \node[above] at (axis cs:2000, 0.935) {\tiny \textcolor{red}{\textbf{Shorter training}}};
      \node[above] at (axis cs:2000, 0.928) {\tiny \textcolor{red}{\textbf{and}}};
      \node[above] at (axis cs:2000, 0.921) {\tiny \textcolor{red}{\textbf{high performance}}};
       
      \addplot[only marks, mark=*, color=red] coordinates {
         (12752, 0.778) 
         (6445, 0.906)  
         (13773, 0.913) 
         (4945, 0.871)  
         (25910, 0.907) 
         (2393, 0.909)  
         (6231, 0.912)  
         (1800, 0.906)  
      };
       
      \node[right] at (axis cs: 12752, 0.778) {\tiny GOGGLE};
      \node[below] at (axis cs: 6445, 0.906) {\tiny STaSy};
      \node[above] at (axis cs: 13773, 0.913) {\tiny GReaT};
      \node[above] at (axis cs: 4945, 0.871) {\tiny CoDi};
      \node[below left] at (axis cs: 25910, 0.907) {\tiny TabDDPM};
      \node[above] at (axis cs: 2393, 0.909) {\tiny TabSyn};
      \node[above] at (axis cs: 6231, 0.912) {\tiny TabDiff};
      \node[below] at (axis cs: 1800, 0.90) {\tiny\bf RecTable (Ours)};
      
      \end{axis}
   \end{tikzpicture}
   }
   \caption{Training time and Machine Learning Efficiency score on the adult dataset. Our proposed method, RecTable, maintains the high performance in downstream task and shorten training time.}
   \label{figure:adult-result}
\end{wrapfigure}
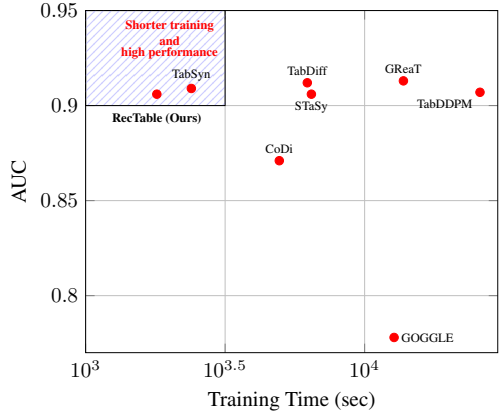

\section{Introduction}
Tabular data is a fundamental data format utilized across various domains, including science, finance~\cite{10.1007/978-3-030-68790-8_50}, medicine~\cite{10.24963/ijcai.2024/670}, healthcare~\cite{doi:10.1504/IJEH.2016.078745}, and e-commerce~\cite{NEDERSTIGT2014296}. Its characteristics encompass aspects such as dataset size, the diversity of categorical features, the distributional properties of numerical data, and the presence or absence of privacy-sensitive information. Tabular data is widely applied in practical scenarios, including data analysis, missing value imputation~\cite{tashiro2021csdi}, data augmentation~\cite{Fonseca2023}, anomaly detection~\cite{shenkar2022anomaly}, and simulation. In machine learning, constructing highly accurate models requires diverse and sufficiently large datasets. However, real-world data is often inaccessible due to challenges such as privacy regulations, high data acquisition costs, data imbalance, and the presence of missing values.

To address these challenges, research has been conducted in recent years on methods for generating high-quality tabular data. Traditionally, data augmentation techniques such as SMOTE~\cite{10.5555/1622407.1622416} and its variants have been widely employed~\cite{Mukherjee_2021,10.1007/11538059_91,10.1007/978-3-642-01307-2_43}. However, these methods struggle to accurately model data distributions and are inherently limited in the diversity of generated data due to their deterministic generation processes. In contrast, the advancement of deep learning has driven substantial progress in tabular data generation through generative models. Early studies introduced approaches based on generative adversarial networks (GANs)~\cite{NIPS2014_f033ed80} and variational autoencoders (VAEs)~\cite{kingma2022autoencodingvariationalbayes}, which enabled more flexible and higher-quality data generation compared to traditional techniques~\cite{1054142,NIPS2017_44a2e080}. More recently, methods leveraging the success of foundation models~\cite{bommasani2022opportunitiesrisksfoundationmodels}, including large language models (LLMs) and diffusion models~\cite{NEURIPS2020_4c5bcfec,pmlr-v37-sohl-dickstein15}, have gained widespread adoption~\cite{pmlr-v202-kotelnikov23a,shi2025tabdiff,zhang2024mixedtype,kim2023stasy,pmlr-v202-lee23i}. These approaches offer significant advantages in learning complex tabular data distributions and generating high-quality synthetic datasets.

However, methods based on LLMs and diffusion models~\cite{pmlr-v37-sohl-dickstein15, NEURIPS2020_4c5bcfec} face significant challenges related to computational costs. LLMs generate data sequentially at the token level, requiring numerous inference steps, which increases generation time. Similarly, diffusion models produce data through an iterative denoising process, necessitating multiple inference steps. These factors not only lead to high computational costs during inference but also demand substantial time and computational resources for model training. Addressing these challenges is crucial for the practical application of tabular data generation.

In this paper, we introduce RecTable, a tabular data generation method that offers faster training compared to methods based on LLMs or diffusion models. RecTable uses rectified flow~\cite{liu2023flow} employed in models such as Stable Diffusion 3\cite{pmlr-v235-esser24a} and Frieren~\cite{wang2024frieren}. It features a simple architecture and an $\ell_2$ loss function, enabling efficient training. To further reduce training time, RecTable avoids transformer-based architectures, as reducing the number of parameters is crucial for computational efficiency. Instead, it incorporates Gated Linear Units (GLU)~\cite{pmlr-v70-dauphin17a, narang-etal-2021-transformer} to enhance accuracy. Additionally, we introduce three modifications to the standard rectified flow training process. Specifically, we adopt the logit-normal timestep distribution~\cite{10.1093/biomet/67.2.261}, as used in Stable Diffusion 3, and adjust the noise distribution to better accommodate mixed-type data.

We demonstrate that RecTable achieves competitive performance compared to diffusion-based methods while outperforming GANs and VAEs. On the adult dataset (containing 32,561 training samples), as shown in \Cref{figure:adult-result}, RecTable achieves shorter training time than existing methods while maintaining the quality of generated data. Furthermore, on some datasets, RecTable surpasses state-of-the-art methods in terms of both the diversity and quality of generated samples.

\section{Related Works}
\subsection{Tabular Data Generation with Deep Learning}
The application of generative models to tabular data is crucial for real-world scenarios. Tabular data is inherently complex, as it consists of both numerical and categorical features. Categorical features are often imbalanced, posing challenges for effective modeling. To address this, CTGAN and TVAE, introduced by \citet{NEURIPS2019_254ed7d2}, use GANs and VAEs, respectively, and remain de facto standards for tabular data generation. GOGGLE~\cite{liu2023goggle} was proposed as an encoder-decoder model based on VAEs. It explicitly captures dependency relationships between columns by representing them as a graph structure. Inspired by advancements in natural language processing, GReaT~\cite{borisov2023language} applies a language modeling approach to tabular data by fine-tuning GPT-2~\cite{radford2019language}, treating each column as a sequence of natural language tokens.

Denoising diffusion models~\cite{pmlr-v37-sohl-dickstein15, NEURIPS2020_4c5bcfec}, which have demonstrated remarkable success in image generation~\cite{dhariwal2021diffusion,saharia2022photorealistic}, have been extensively studied for tabular data generation. Several methods, including TabDDPM~\cite{pmlr-v202-kotelnikov23a}, STaSy~\cite{kim2023stasy}, CoDi~\cite{pmlr-v202-lee23i}, CDTD~\cite{mueller2025continuous}, and TabDiff~\cite{shi2025tabdiff}, apply diffusion models directly in data space.

In image generation, Latent Diffusion Models~\cite{Rombach_2022_CVPR} have emerged as the mainstream approach, significantly reducing computational complexity by projecting data into a latent space. TabSyn~\cite{zhang2024mixedtype} extends this idea to tabular data, achieving comparable performance to conventional diffusion models while requiring lower computational cost.

\subsection{Rectified Flow and Its Application}
Diffusion models, which transform distributions using stochastic differential equations, are known to be inefficient for both training and generation. In contrast, Flow Matching~\cite{lipman2023flow}, an extension of Continuous Normalizing Flow~\cite{NEURIPS2018_69386f6b}, employs ordinary differential equations for distribution transformation, offering a more efficient alternative. Flow Matching, which learns vector fields to map between distributions, has been improved in various ways within the framework of optimal transport. Among these advancements, Rectified Flow~\cite{albergo2023building,liu2023flow}, which utilizes linear interpolations, is particularly well-suited for high-dimensional distributions such as images due to its simplicity and scalability. Notably, such as Stable Diffusion 3~\cite{pmlr-v235-esser24a} and InstaFlow~\cite{liu2024instaflow} use rectified flow to enable faster generation while maintaining or even improving output quality. Beyond image generation, the application of rectified flow is being actively explored in other domains, including video generation~\cite{wang2024frieren}, audio generation~\cite{10447822,10445948}, and language modeling~\cite{zhang-etal-2024-languageflow}.

\section{Method}
\subsection{Preliminaries: Rectified Flow}
Rectified Flow learns a transport map $T:\mathbb{R}^d\to\mathbb{R}^d$ that transforms a data point $z_0\in\mathbb{R}^d$ sampled from the source distribution $\pi_0$ to a corresponding data point $z_1\in\mathbb{R}^d$ sampled from the target distribution $\pi_1$. In generative modeling, $\pi_0$ is typically a known noise distribution—commonly a gaussian distribution $\mathcal{N}(0, 1)$—while $\pi_1$ represents the unknown data distribution to be learned.  

Rectified Flow trains a velocity field $v_{\theta}$, parameterized by a deep neural network, using the following loss function:

\begin{align}
\mathcal{L}=\mathbb{E}_{(z_0, z_1)\sim(\pi_0, \pi_1), t\sim p_t}[\|v_{\theta}(z_t, t) - (z_1-z_0)\|_2^2]
\end{align}

where $z_t$ is an interpolated state defined as $z_t=tz_1+(1-t)z_0$, and $p_t$ denotes the timestep distribution. To generate samples, we solve the following ordinary differential equation from $t=1$ to $t=0$ 

\begin{align}
   \mathrm{d}z_t=v_{\theta}(z_t, t)\mathrm{d}t.
\end{align}

\subsection{Network Architecture}
We replace the MLP blocks in the TabDDPM architecture~\cite{pmlr-v202-kotelnikov23a} with Gated Linear Unit (GLU) blocks~\cite{pmlr-v70-dauphin17a, narang-etal-2021-transformer} to better capture nonlinear relationships between features. The GLU block used in RecTable is described as:

\begin{align}
\mathtt{GLU}(\boldsymbol{x})=\mathrm{Dropout}\left(\left(\boldsymbol{x}\boldsymbol{W}_1+\boldsymbol{b}_1\right)\otimes\mathrm{Sigmoid}\left(\boldsymbol{x}\boldsymbol{W}_2+\boldsymbol{b}_2\right)\right)
\end{align}

where $\boldsymbol{W}_1, \boldsymbol{W}_2$ are the weight matrices of the linear layers, and $\otimes$ denotes the Hadamard product (element-wise product). 

Compared to the attention mechanism~\cite{NIPS2017_3f5ee243}, GLU requires fewer parameters, which contributes to a reduction in training time. Following TabDDPM, we embed the timestep information using a sinusoidal time embedding, as proposed in~\citet{NEURIPS2020_4c5bcfec} and \citet{dhariwal2021diffusion}.

\begin{align}
t\_emb = \mathtt{Linear}(\mathtt{SiLU}(\mathtt{Linear}(\mathtt{SinTimeEmb}(t))))
\end{align}

All linear layers in timestep embedding have 128 dimensions.

\subsection{Training Strategies}
We introduce three modifications to the standard training strategies of rectified flow.

\subsubsection{Timestep Distribution}
In Stable Diffusion 3~\cite{pmlr-v235-esser24a}, various timestep distribution settings are considered. We use the best timestep distribution in their experiments, logit-normal distribution~\cite{10.1093/biomet/67.2.261}. It is represented as

\begin{align}
\pi_{\mathrm{ln}}(t;m, s)=\dfrac{1}{s\sqrt{2\pi}}\dfrac{1}{t(1-t)}\exp\left(-\dfrac{(\mathrm{logit}(t)-m)^2}{2s^2}\right)
\end{align}

where $\mathrm{logit}(t)=\log\dfrac{t}{1-t}$, $m$ and $s$ are hyperparameters respectively. We use $m=0, s=1$. This is the best setting in their experiments.

\subsubsection{Noise Distribution}
In general, a gaussian distribution is commonly used as the distribution $\pi_0$. However, tabular data consists of mixed-type features: numerical and categorical. While numerical data typically follows a gaussian distribution, categorical data follows a categorical distribution. Let $x=[x_{\mathrm{num}}, x_{\mathrm{cat}_1}, \ldots,x_{\mathrm{cat}_{C}}]$ be a tabular data sample with $N_{\mathrm{num}}$ numerical features and $C$ catgeorical features. Each categorical feature $x_{\mathrm{cat}_i}$ has $K_i$ cardinalities and is represented using a one-hot encoding $x_{\mathrm{cat}_i}^{\mathrm{ohe}}\in\{0, 1\}^{K_i}$. We assume that the numerical features follow a gaussian distribution, $x_{\mathrm{num}}\sim\mathcal{N}(0, 1)$, while the categorical features follow categorical distributions, $x_{\mathrm{cat}_i}\sim\mathrm{Cat}_i$. To match these distributions during training, we adopt a hybrid noise model: a gaussian distribution for numerical features and a uniform distribution for categorical features. The initial noise $z_0\sim\pi_0$ is thus defined as:

\begin{align}
z_0=[z_{\mathrm{num}}, z_{\mathrm{cat}_1}, \ldots, z_{\mathrm{cat}_C}],\quad z_{\mathrm{num}}\sim\mathcal{N}(0, 1),\quad z_{\mathrm{cat}_i}\sim\mathcal{U}(1, K_i)
\end{align}

\subsubsection{Reflow}
Reflow~\cite{liu2023flow} is a rectified flow process that utilizes real noise and corresponding synthetic data. This process generates ODE straight trajectories after being repeated $k$ times to enable fast generation. However, maintaining strict trajectory straightness is not a prerequisite for fast generation~\cite{wang2025rectified}. Based on this insight, we do not execute the reflow process.

\section{Experiments}
\label{sec:exp}
\subsection{Experimental Settings}
\paragraph{Datasets.} Following ~\cite{zhang2024mixedtype}, we use six real-world tabular datasets: Adult~\cite{adult_2}, Default~\cite{default_of_credit_card_clients_350}, Shoppers~\cite{online_shoppers_purchasing_intention_dataset_468}, Magic~\cite{magic_gamma_telescope_159}, Beijing~\cite{beijing_pm2.5_381}, and News~\cite{online_news_popularity_332}. Each dataset has numerical and categorical column. More details are shown in \Cref{appendix:datasets}.

\paragraph{Baselines.} We compare the proposed RecTable with nine methods. GAN-based model: CTGAN~\cite{NEURIPS2019_254ed7d2}. VAE-based models: TVAE~\cite{NEURIPS2019_254ed7d2} and GOGGLE~\cite{liu2023goggle}. Autoregressive language model: GReaT~\cite{borisov2023language}. Diffusion-based models: TabDDPM~\cite{pmlr-v202-kotelnikov23a}, CoDi~\cite{pmlr-v202-lee23i}, STaSy~\cite{kim2023stasy}, TabSyn~\cite{zhang2024mixedtype}, and TabDiff~\cite{shi2025tabdiff}.

\paragraph{Evaluation Methods.} Following previous studies~\cite{zhang2024mixedtype,shi2025tabdiff}, We evaluate the generated data from two aspects. 1) \textbf{Fidelity}: Shape, Trend, detection score (C2ST)~\cite{lopez-paz2017revisiting}, $\alpha$-Precision~\cite{pmlr-v162-alaa22a} and $\beta$-Recall~\cite{pmlr-v162-alaa22a} assess how well the generated data can faithfully recover the ground-truth data distribution. 2): \textbf{Machine Learning Efficiency} (MLE): To evaluate the utility of the generated data for downstream machine learning tasks, we first split the real data into training, validation, and test set. The generative models are trained on the training set and then used to generate synthetic data of the same size as the original training set. We train a classification or regression model using XGBoost~\cite{10.1145/2939672.2939785} on the synthetic data and evaluate their performance on the test set. MLE performance is measured using the AUC score for classification tasks and RMSE for regression tasks. We report the mean and standard deviation of the AUC and RMSE score over 20 independent experiments.

\subsection{Implementation Details}
We implement RecTable using PyTorch 2.3.1~\cite{NEURIPS2019_bdbca288}. RecTable has four GLU Blocks and one MLP head, the hidden sizes are 1024, 2048, 1024, and 1024, respectively. The model is optimized using Adam optimizer~\cite{kingma2017adammethodstochasticoptimization} with a learning rate of $2\times10^{-4}$ and $\beta_1=0.9$. RecTable is trained for 30,000 iterations with a batch size of 4096 using the $\ell_2$ loss function. All experiments except GReaT is conducted on an NVIDIA RTX A5000 GPU with 24GB memory (GReaT is trained on four NVIDIA RTX A5000 GPUs). To satisfy $x_{\mathrm{num}}\sim\mathcal{N}(0, 1)$, we use QuantileTransformer\footnote{\url{https://scikit-learn.org/stable/modules/generated/sklearn.preprocessing.QuantileTransformer.html}} from Scikit-learn~\cite{JMLR:v12:pedregosa11a} for numerical features as preprocessing. For one-hot encoding categorical features, we also use scikit-learn. We use the Runge-Kutta method of order 5(4) from Scipy~\cite{Virtanen2020} for generation following \citet{liu2023flow}. 

\subsection{Results of Data Fidelity}
The results for Shape and Trend are presented in Tables~\ref{table:exp-shape} and \ref{table:exp-trend}, respectively. In both metrics, RecTable performs competitively, ranking just below TabDiff and TabSyn. Since RecTable lags behind state-of-the-art methods in Shape, we consider that capturing complex column-to-column relationships remains a challenge. Notably, when compared to TabDDPM—which shares a similar architecture—TabDDPM's simple MLP-based design excels on low-dimensional datasets. However, as dimensionality increases (e.g., in the news dataset), the performance of MLP-based models declines, whereas RecTable's GLU-based architecture effectively mitigates this deterioration. In contrast to Shape, RecTable achieves state-of-the-art performance on certain datasets in Trend evaluation as shown in \Cref{table:exp-trend}. Similar to Shape, it also demonstrates robustness against performance degradation in high-dimensional datasets such as the news. Moreover, a comparison with \Cref{table:exp-mle} indicates that the scores in Tables~\ref{table:exp-shape} and \ref{table:exp-trend} do not always directly correlate with downstream task performance.

\begin{table}[!t] 
   \centering
   \caption{Performance comparison on the error rates (\%) of Shape.} 
   \label{table:exp-shape}
   \resizebox{\columnwidth}{!}{
   \begin{threeparttable}{
   \begin{tabular}{lcccccc|c}
      \toprule[0.8pt]
      Method & Adult & Default & Shoppers & Magic & Beijing & News & Average \\
      \midrule 
      CTGAN$^{1}$ & $16.84${\tiny$\pm$ $0.03$} & $16.83${\tiny$\pm$$0.04$} & $21.15${\tiny$\pm0.10$} & $9.81${\tiny$\pm0.08$} & $21.39${\tiny$\pm0.05$} & $16.09${\tiny$\pm 0.02$} & $17.02$  \\
      TVAE$^{1}$ & $14.22${\tiny$\pm0.08$} & $10.17${\tiny$\pm$$0.05$} & $24.51${\tiny$\pm0.06$} & $8.25${\tiny$\pm0.06$} & $19.16${\tiny$\pm0.06$} & $16.62${\tiny$\pm0.03$} & $15.49$  \\
      GOGGLE$^{1}$ & $16.97$ & $17.02$ & $22.33$ & $1.90$ & $16.93 $ & $25.32$ & $16.75$ \\
      GReaT$^{1}$ & $12.12${\tiny$\pm$$0.04$} & $19.94${\tiny$\pm$$0.06$} & $14.51${\tiny$\pm0.12$} & $16.16${\tiny$\pm0.09$} & $8.25${\tiny$\pm0.12$} & N/A$^{2}$ & $14.20$ \\
      STaSy$^{1}$ & $11.29${\tiny$\pm0.06$} & $5.77${\tiny$\pm0.06$} & $9.37${\tiny$\pm0.09$} & $6.29${\tiny$\pm0.13$} & $6.71${\tiny$\pm0.03$} & $6.89${\tiny$\pm0.03$} & $7.72$ \\
      CoDi$^{1}$ & $21.38${\tiny$\pm0.06$} & $15.77${\tiny$\pm$ $0.07$} & $31.84${\tiny$\pm0.05$} & $11.56${\tiny$\pm0.26$} & $16.94${\tiny$\pm0.02$} & $32.27${\tiny$\pm0.04$} & $21.63$ \\
      TabDDPM$^{1}$ & $1.75${\tiny$\pm0.03$} & $1.57${\tiny$\pm$ $0.08$} & $2.72${\tiny$\pm0.13$} & $1.01${\tiny$\pm0.09$} & $1.30${\tiny$\pm0.03$} & $78.75${\tiny$\pm0.01$} & $14.53$ \\
      TabSyn$^{1}$ & ${0.81}${\tiny${\pm0.05}$} & \textbf{1.01}{\tiny$\bm\pm$\textbf{0.08}} & ${1.44}${\tiny${\pm0.07}$} & ${1.03}${\tiny${\pm0.14}$} & ${1.26}${\tiny${\pm0.05}$} & \textbf{2.06}{\tiny$\bm\pm$\textbf{0.04}} & $1.27$ \\
      TabDiff$^{1}$ & \textbf{0.63}{\tiny$\bm\pm$\textbf{0.05}} & ${1.24}${\tiny${\pm0.07}$} & \textbf{1.28}{\tiny$\bm\pm$\textbf{0.09}} & \textbf{0.78}{\tiny$\bm\pm$\textbf{0.08}} & \textbf{1.03}{\tiny$\bm\pm$\textbf{0.05}} & ${2.35}${\tiny${\pm0.03}$} & \textbf{1.22} \\
      \midrule
      RecTable & $3.63${\tiny${\pm0.07}$} & $1.74${\tiny${\pm0.04}$} & $4.44${\tiny${\pm0.17}$} & $1.21${\tiny${\pm0.12}$} & $5.42${\tiny${\pm0.10}$} & $6.28${\tiny${\pm0.44}$} & $3.79$ \\
   \bottomrule[1.0pt] 
   \end{tabular}
   }
   \begin{tablenotes}
   \item[1] The results of all baselines are taken from \citet{shi2025tabdiff}. 
   \item[2] The N/A represents the results are not provided in \citet{shi2025tabdiff} and \citet{zhang2024mixedtype}due to out-of-memory.
   \end{tablenotes}
   \vspace{-5pt}
   \end{threeparttable}
   }
\end{table}

\begin{table}[!t] 
   \centering
   \caption{Performance comparison on the error rates (\%) of Trend.} 
   \label{table:exp-trend}
   \resizebox{\columnwidth}{!}{
   \begin{threeparttable}{
   \begin{tabular}{lcccccc|c}
      \toprule[0.8pt]
      Method & Adult & Default & Shoppers & Magic & Beijing & News & Average \\
      \midrule 
      CTGAN$^{1}$ & $20.23${\tiny$\pm1.20$} & $26.95${\tiny$\pm0.93$} & $13.08${\tiny$\pm0.16$} & $7.00${\tiny$\pm0.19$} & $22.95${\tiny$\pm0.08$} & $5.37${\tiny$\pm0.05$} & $15.93$ \\
      TVAE$^{1}$ & $14.15${\tiny$\pm0.88$} & $19.50${\tiny$\pm$$0.95$} & $18.67${\tiny$\pm0.38$} & $5.82${\tiny$\pm0.49$} & $18.01${\tiny$\pm0.08$}  & $6.17${\tiny$\pm0.09$} & $13.72$ \\
      GOGGLE$^{1}$ & $45.29$ & $21.94$ & $23.90$ & $9.47$ & $45.94$ & $23.19$ & $28.29$ \\
      GReaT$^{1}$ & $17.59${\tiny$\pm0.22$} & $70.02${\tiny$\pm$$0.12$} & $45.16${\tiny$\pm0.18$} & $10.23${\tiny$\pm0.40$} & $59.60${\tiny$\pm0.55$} & N/A$^{2}$ & $44.24$ \\
      STaSy$^{1}$ & $14.51${\tiny$\pm0.25$} & $5.96${\tiny$\pm$$0.26$} & $8.49${\tiny$\pm0.15$} & $6.61${\tiny$\pm0.53$} & $8.00${\tiny$\pm0.10$} & $3.07${\tiny$\pm0.04$} & $7.77$ \\
      CoDi$^{1}$ & $22.49${\tiny$\pm0.08$} & $68.41${\tiny$\pm$$0.05$} & $17.78${\tiny$\pm0.11$} & $6.53${\tiny$\pm0.25$} & $7.07${\tiny$\pm0.15$} & $11.10${\tiny$\pm0.01$} & $23.23$ \\ 
      TabDDPM$^{1}$ & $3.01${\tiny$\pm0.25$} & $4.89${\tiny$\pm0.10$} & $6.61${\tiny$\pm0.16$} & $1.70${\tiny$\pm0.22$} & $2.71${\tiny$\pm0.09$} & $13.16${\tiny$\pm0.11$} & $5.35$ \\
      TabSyn$^{1}$ & ${1.93}${\tiny${\pm0.07}$} & ${2.81}${\tiny${\pm0.48}$} & ${2.13}${\tiny${\pm0.10}$} & ${0.88}${\tiny${\pm0.18}$} & ${3.13}${\tiny${\pm0.34}$}  & ${1.52}${\tiny${\pm0.03}$} & ${2.07}$ \\
      TabDiff$^{1}$ & \textbf{1.49}{\tiny$\bm\pm$\textbf{0.16}} & $2.55${\tiny$\pm0.75$} & \textbf{1.74}{\tiny$\bm\pm$\textbf{0.08}} & \textbf{0.76}{\tiny$\bm\pm$\textbf{0.12}} & $2.59${\tiny$\pm0.15$} & \textbf{1.28}{\tiny$\bm\pm$\textbf{0.04}} & \textbf{1.73} \\
      \midrule
      RecTable & $6.04${\tiny$\pm0.07$} & \textbf{1.13}{\tiny$\bm\pm$\textbf{0.03}} & $3.51${\tiny$\pm0.37$} & $1.80${\tiny$\pm0.55$} & \textbf{2.41}{\tiny$\bm\pm$\textbf{0.07}} & $2.03${\tiny$\pm0.19$} & $2.82$ \\
   \bottomrule[1.0pt] 
   \end{tabular}
   }
   \begin{tablenotes}
      \item[1] The results of all baselines are taken from \citet{shi2025tabdiff}. 
      \item[2] The N/A represents the results are not provided in \citet{shi2025tabdiff} and \citet{zhang2024mixedtype} due to out-of-memory.
   \end{tablenotes}
   \vspace{-5pt}
   \end{threeparttable}
}
\end{table}

Following \citet{zhang2024mixedtype} and \citet{shi2025tabdiff}, we calculate $\alpha$-Precision, which measures generation quality, and $\beta$-Recall, which measures how well the synthetic data covers the real data distribution. Tables~\ref{table:exp-alpha-precision} and \ref{table:exp-beta-recall} are presented the results of $\alpha$-Precision and $\beta$-Recall, respectively. We observe a similar trend to Shape results in \Cref{table:exp-alpha-precision}. However, RecTable achives the highest $\beta$-Recall scores in \Cref{table:exp-beta-recall}, indicating that it can generate diverse samples. In the news dataset, which TabDDPM cannot generate meaningful samples according to \citet{zhang2024mixedtype}, RecTable successfully generates meaningful samples with high-quality. We consider that GLU contributes to complement nonlinear relationships between features.

\begin{table}[!ht] 
   \centering
   \caption{Comparison of $\alpha$-Precision scores. Higher scores indicate better performance.}  
   \label{table:exp-alpha-precision}
   \resizebox{\columnwidth}{!}{
      \begin{threeparttable}{
       \begin{tabular}{lcccccc|cc}
           \toprule[0.8pt]
           Methods & Adult & Default & Shoppers & Magic & Beijing & News & Average & Ranking \\
           \midrule 
           CTGAN$^{1}$ & $77.74${\tiny$\pm0.15$} & $62.08${\tiny$\pm0.08$} & $76.97${\tiny$\pm0.39$} & $86.90${\tiny$\pm0.22$} & $96.27${\tiny$\pm0.14$} & $96.96${\tiny$\pm0.17$} & $82.82$ & $6$ \\
           TVAE$^{1}$     & $98.17${\tiny$\pm0.17$}  & $85.57${\tiny$\pm0.34$} & $58.19${\tiny$\pm0.26$} & $86.19${\tiny$\pm0.48$} & $97.20${\tiny$\pm0.10$} & $86.41${\tiny$\pm0.17$} & $85.29$  & $8$ \\
           GOGGLE$^{1}$  & $50.68$  & $68.89$ & $86.95$ & $90.88$ & $88.81$ & $86.41$ & $78.77$ & $10$ \\
           GReaT$^{1}$    & $55.79${\tiny$\pm0.03$}  & $85.90${\tiny$\pm0.17$}  & $78.88${\tiny$\pm0.13$} & $85.46${\tiny$\pm0.54$} & \textbf{98.32}{\tiny$\bm\pm$\textbf{0.22}} & N/A$^2$ & $80.87$ & $7$ \\
           STaSy$^{1}$    & $82.87${\tiny$\pm0.26$} & $90.48${\tiny$\pm0.11$} & $89.65${\tiny$\pm0.25$} & $86.56${\tiny$\pm0.19$} & $89.16${\tiny$\pm0.12$} & $94.76${\tiny$\pm0.33$} & $88.91$ & $4$ \\
           CoDi$^{1}$ & $77.58${\tiny$\pm0.45$} & $82.38${\tiny$\pm0.15$}  & $94.95${\tiny$\pm0.35$} & $85.01${\tiny$\pm0.36$} & ${98.13}${\tiny${\pm0.38}$} & $87.15${\tiny$\pm0.12$} & $87.53$ & $5$ \\
           TabDDPM$^{1}$  & $96.36${\tiny$\pm0.20$}  & $97.59${\tiny$\pm0.36$} & $88.55${\tiny$\pm0.68$} & $98.59${\tiny$\pm0.17$} & $97.93$\tiny${\pm0.30}$ & $0.00${\tiny$\pm0.00$} & $79.84$ & $9$ \\
           TabSyn$^{1}$ & \textbf{99.39}{\tiny$\bm\pm$\textbf{0.18}}  & $98.65$\tiny$\pm0.23$ & ${98.36}$\tiny${\pm0.52}$ & ${99.42}$\tiny${\pm0.28}$ & ${97.51}$\tiny${\pm0.24}$& ${95.05}$\tiny${\pm0.30}$ & 98.06 & $2$ \\
           TabDiff$^{1}$ & ${99.02}$\tiny${\pm0.20}$ & \textbf{98.49}{\tiny$\bm\pm$\textbf{0.28}} & \textbf{99.11}{\tiny$\bm\pm$\textbf{0.34}} & \textbf{99.47}{\tiny$\bm\pm$\textbf{0.21}} & ${98.06}$\tiny${\pm0.24}$ & \textbf{97.36}{\tiny$\bm\pm$\textbf{0.17}} & \textbf{98.59} & $1$ \\
           \midrule
           RecTable & 86.70{\tiny$\pm$0.26} & 97.61{\tiny$\pm$0.27} & 93.24{\tiny$\pm$0.61} & 99.20{\tiny$\pm$0.22} & 97.96{\tiny$\pm$0.26} & 95.36{\tiny$\pm$0.27} & 95.01 & 3 \\
     \bottomrule[1.0pt] 
     \end{tabular}
      }
     \begin{tablenotes}
      \item[1] The results of all baselines are taken from \citet{shi2025tabdiff}. 
      \item[2] The N/A represents the results are not provided in \citet{shi2025tabdiff} and \citet{zhang2024mixedtype}due to out-of-memory.
   \end{tablenotes}
   \vspace{-5pt}
   \end{threeparttable}
 }
\end{table}

\begin{table}[!ht] 
   \centering
   \caption{Comparison of $\beta$-Recall scores. Higher scores indicate better results.}  
   \label{table:exp-beta-recall}
   \resizebox{\columnwidth}{!}{
      \begin{threeparttable}{
      \begin{tabular}{lcccccc|cc}
         \toprule[0.8pt]
         Methods & Adult & Default & Shoppers & Magic & Beijing & News & Average & Ranking \\
         \midrule 
         CTGAN$^{1}$ & $30.80${\tiny$\pm0.20$} & $18.22${\tiny$\pm0.17$} & $31.80${\tiny$\pm0.350$} & $11.75${\tiny$\pm0.20$} & $34.80${\tiny$\pm0.10$} & $24.97${\tiny$\pm0.29$} & $25.39$ & $9$ \\
         TVAE$^{1}$ & $38.87${\tiny$\pm0.31$} & $23.13${\tiny$\pm0.11$} & $19.78${\tiny$\pm0.10$} & $32.44${\tiny$\pm0.35$} & $28.45${\tiny$\pm0.08$} & $29.66${\tiny$\pm0.21$} & $28.72$  & $8$ \\
         GOGGLE$^{1}$  & $8.80$  & $14.38$ & $9.79$ & $9.88$ & $19.87$ & $2.03$ & $10.79$ & $10$ \\
         GReaT$^{1}$ & ${49.12}${\tiny${\pm0.18}$} & $42.04${\tiny$\pm0.19$}  & $44.90${\tiny$\pm0.17$} & $34.91${\tiny$\pm0.28$} & $43.34${\tiny$\pm0.31$} & N/A$^{2}$ & $43.34$ & $4$ \\
         STaSy$^{1}$ & $29.21${\tiny$\pm0.34$} & $39.31${\tiny$\pm0.39$} & $37.24${\tiny$\pm0.45$} & ${53.97}${\tiny${\pm0.57}$} & $54.79${\tiny$\pm0.18$} & $39.42${\tiny$\pm0.32$} & $42.32$ & $5$ \\
         CoDi$^{1}$ & $9.20${\tiny$\pm0.15$} & $19.94${\tiny$\pm0.22$} & $20.82${\tiny$\pm0.23$} & $50.56${\tiny$\pm0.31$} & $52.19${\tiny$\pm0.12$} & $34.40${\tiny$\pm0.31$} & $38.19$ & $7$ \\
         TabDDPM$^{1}$ & $47.05${\tiny$\pm0.25$}  & $47.83${\tiny$\pm0.35$} & $47.79${\tiny$\pm0.25$} & $48.46${\tiny$\pm0.42$} & ${56.92}$\tiny${\pm0.13}$ & $0.00${\tiny$\pm0.00$} & $41.34$ & $6$  \\
         TabSyn$^{1}$ & ${47.92}$\tiny${\pm0.23}$ & ${46.45}$\tiny${\pm0.35}$ & ${49.10}$\tiny${\pm0.60}$ & ${48.03}$\tiny${\pm0.50}$ & $59.15$\tiny${\pm0.22}$ & $43.01$\tiny$\pm0.28$ & ${48.94}$ & $3$ \\
         TabDiff$^{1}$ & \textbf{51.64}{\tiny$\bm\pm$\textbf{0.20}}  & \textbf{51.09}{\tiny$\bm\pm$\textbf{0.25}} & $49.75$\tiny$\pm0.64$ & ${48.01}$\tiny${\pm0.31}$ & \textbf{59.63}{\tiny$\bm\pm$\textbf{0.23}} & ${42.10}$\tiny${\pm0.32}$ & $50.37$ & $2$ \\
         \midrule
         RecTable & 41.75{\tiny$\pm$0.27} & 48.04{\tiny$\pm$0.46} & \textbf{53.72}{\tiny$\bm\pm$\textbf{0.49}} & \textbf{57.71}{\tiny$\bm\pm$\textbf{0.33}} & 54.70{\tiny$\pm$0.25} & \textbf{57.55}{\tiny$\bm\pm$\textbf{0.22}} & \textbf{52.25} & 1 \\
      \bottomrule[1.0pt] 
      \end{tabular}
      }
      \begin{tablenotes}
         \item[1] The results of all baselines are taken from \citet{shi2025tabdiff}. 
         \item[2] The N/A represents the results are not provided in \citet{shi2025tabdiff} and \citet{zhang2024mixedtype}due to out-of-memory.
      \end{tablenotes}
      \vspace{-5pt}
      \end{threeparttable}
 }
\end{table}

We further investigate the similarity between synthetic and real data using C2ST test~\cite{lopez-paz2017revisiting}, which measures the difficulty of distinguishing synthetic data from real data. We present the results in \Cref{table:exp-c2st}. We observe a similar trend to the Shape and the Trend results. We believe that C2ST scores are due to the poor Shape score and the failure of the generated data to reproduce the relationships between columns in the real data.

\begin{table}[!ht]  
   \centering
   \caption{Detection score (C2ST) using logistic regression classifier. Higher scores indicate superior performance.} 
   \label{table:exp-c2st}
   \resizebox{\columnwidth}{!}{
      \begin{threeparttable}{
       \begin{tabular}{lcccccc|cc}
           \toprule[0.8pt]
           Method & Adult & Default & Shoppers & Magic & Beijing & News & Average & Ranking \\
           \midrule
           CTGAN$^{1}$   & $0.5949$ & $0.4875$ & $0.7488$ & $0.6728$ & $0.7531$ & $0.6947$ & $0.6586$ & 5 \\
           TVAE$^{1}$    & $0.6315$ & $0.6547$ & $0.2962$ & $0.7706$ & $0.8659$ & $0.4076$ & $0.6044$ & 7 \\
           GOGGLE$^{1}$  & $0.1114$ & $0.5163$ & $0.1418$ & $0.9526$ & $0.4779$ & $0.0745$ & $0.3791$ & 10 \\
           GReaT$^{1}$   & $0.5376$ & $0.4710$ & $0.4285$ & $0.4326$ & $0.6893$ & N/A$^2$  & $0.5118$ & 8 \\
           STaSy$^{1}$   & $0.4054$ & $0.6814$ & $0.5482$ & $0.6939$ & $0.7922$ & $0.5287$ & $0.6083$ & 6 \\
           CoDi$^{1}$    & $0.2077$ & $0.4595$ & $0.2784$ & $0.7206$ & $0.7177$ & $0.0201$ & $0.4007$ & 9 \\
           TabDDPM$^{1}$ & $0.9755$ & $0.9712$ & $0.8349$ & \textbf{0.9998} & $0.9513$ & $0.0002$ & $0.7888$ & 4 \\
           TabSyn$^{1}$  & $0.9910$ & \textbf{0.9826} & $0.9662$ & $0.9960$ & $0.9528$ & $0.9255$ & $0.9690$ & 2 \\
           TabDiff$^{1}$ & \textbf{0.9950} & $0.9774$ & \textbf{0.9843} & $0.9989$ & \textbf{0.9781} & \textbf{0.9308} & \textbf{0.9774} & 1 \\
           \midrule
           RecTable      & $0.8076$ & $0.9057$ & $0.8358$ & $0.9726$ & $0.7152$ & $0.7829$ & $0.8366$ & 3 \\
     \bottomrule[1.0pt] 
     \end{tabular}
   }
   \begin{tablenotes}
      \item[1] The results of all baselines are taken from \citet{shi2025tabdiff}. 
      \item[2] The N/A represents the results are not provided in \citet{shi2025tabdiff} and \citet{zhang2024mixedtype}due to out-of-memory.
   \end{tablenotes}
   \vspace{-5pt}
   \end{threeparttable}
   }
\end{table}

\subsection{Results of Machine Learning Efficiency}
We present the MLE results for each dataset in \Cref{table:exp-mle}. In this evaluation, an XGBoost classifier or regressor is trained using the generated samples, and its performance is measured on the test data. If the quality of the generated data is close to that of real data, the scores of the classifier or regressor is expected to achive high scores. From the results in \Cref{table:exp-mle}, RecTable achieves state-of-the-art performance on half of the datasets, which indicates that high-quality generation is feasible. Overall, its performance is comparable to that of the state-of-the-art method, TabDiff. Even for the datasets where the scores are low in Tables~\ref{table:exp-shape} and \ref{table:exp-trend}, MLE demonstrates high performance. Remarkably, in the news dataset, RecTable achives results that surpass those obtained with real data.

\begin{table}[!t] 
   \centering
   \caption{Evaluation of MLE (Machine Learning Efficiency): AUC and RMSE are used for classification and regression tasks, respectively.}
   \label{table:exp-mle}
   \resizebox{\columnwidth}{!}{
   \begin{threeparttable}{
   \begin{tabular}{lcccccc|c}
      \toprule[0.8pt]
      \multirow{2}{*}{Methods} & Adult & Default & Shoppers & Magic & Beijing & News & Average Gap \\
      \cmidrule{2-8} 
      & AUC $\uparrow$ & AUC $\uparrow$ & AUC $\uparrow$ & AUC $\uparrow$ & RMSE $\downarrow$ & RMSE $\downarrow$ & $\%$ \\
      \midrule 
      Real$^{1}$ & $.927${\tiny$\pm.000$} & $.770${\tiny$\pm.005$} & $.926${\tiny$\pm.001$} & $.946${\tiny$\pm.001$} & $.423${\tiny$\pm.003$} & $.842${\tiny$\pm.002$} & $0.0$ \\
      \midrule
      CTGAN$^{1}$ & $.886${\tiny$\pm.002$} &$.696${\tiny$\pm.005$} & $.875${\tiny$\pm.009$} & $.855${\tiny$\pm.006$} & $.902${\tiny$\pm.019$} & $.880${\tiny$\pm.016$} & $24.5$ \\
      TVAE$^{1}$ & $.878${\tiny$\pm.004$} & $.724 ${\tiny$\pm.005$} & $.871${\tiny$\pm.006$} & $.887${\tiny$\pm.003$} & $.770${\tiny$\pm.011$} & $1.01${\tiny$\pm.016$} & $20.9$ \\
      GOGGLE$^{1}$ & $.778${\tiny$\pm.012$} & $.584${\tiny$\pm.005$} & $.658${\tiny$\pm.052$} & $.654${\tiny$\pm.024$} & $1.09${\tiny$\pm.025$} & $.877${\tiny$\pm.002$} & $43.6$ \\
      GReaT$^{1}$ & \textbf{.913}{\tiny$\bm\pm$\textbf{.003}} &$.755${\tiny$\pm.006$} & $.902${\tiny$\pm.005$} & $.888${\tiny$\pm.008$} & $.653${\tiny$\pm.013$} & N/A$^{2}$ & $13.3$ \\
      STaSy$^{1}$  & $.906${\tiny$\pm.001$} & $.752${\tiny$\pm.006$} & $.914${\tiny$\pm.005$} & $.934${\tiny$\pm.003$} & $.656${\tiny$\pm.014$} & $.871${\tiny$\pm.002$} & $10.9$ \\ 
      CoDi$^{1}$ & $.871${\tiny$\pm.006$} & $.525${\tiny$\pm.006$} & $.865${\tiny$\pm.006$} & $.932${\tiny$\pm.003$} & $.818${\tiny$\pm.021$} & $1.21${\tiny$\pm.005$} & $30.5$ \\
      TabDDPM$^{1}$ & $.907${\tiny$\pm.001$}  & $.758${\tiny$\pm.004$}& $.918${\tiny$\pm.005$} & $.935${\tiny$\pm.003$} & $.592${\tiny$\pm.011$}& $4.86${\tiny$\pm3.04$} & $9.14$ \\
      TabSyn$^{1}$ & $.909${\tiny${\pm.001}$} & \textbf{.763}{\tiny$\bm\pm$\textbf{.002}} & $.914${\tiny${\pm.004}$} & $.937${\tiny$\pm.002$} & ${.580}${\tiny$\pm.009$} & $.862${\tiny$\pm.024$} & ${7.43}$ \\
      TabDiff$^{1}$ & .912{\tiny$\pm$.002} & \textbf{.763}{\tiny$\bm\pm$\textbf{.005}} & \textbf{.921}{\tiny$\bm\pm$\textbf{.004}} & $.936${\tiny${\pm.003}$} & \textbf{.555}{\tiny$\bm\pm$\textbf{.013}} & ${.866}${\tiny${\pm.021}$} & \textbf{6.36} \\
      \midrule 
      RecTable & .906{\tiny$\pm$.002} & .754{\tiny$\pm$.004} & $.904${\tiny$\pm$.010} & \textbf{.939}{\tiny$\bm\pm$\textbf{.003}} & \textbf{.555}{\tiny$\bm\pm$\textbf{.011}} & \textbf{.840}{\tiny$\bm\pm$\textbf{.004}} & $6.40$ \\
   \bottomrule[1.0pt] 
   \end{tabular}
   }
   \begin{tablenotes}
      \item[1] The results of all baselines are taken from \citet{shi2025tabdiff}. 
      \item[2] The N/A represents the results are not provided in \citet{shi2025tabdiff} and \citet{zhang2024mixedtype}due to out-of-memory.
   \end{tablenotes}
   \vspace{-5pt}
   \end{threeparttable}
   }
\end{table}

\subsection{Training Time}
We present the measured training time on the adult dataset in \Cref{table:exp-time}. RecTable achieves the fastest training time while maintaining high-quality data generation. Although its performance is competitive with state-of-the-art methods, it has the shortest training among them. We attribute this efficiency to RecTable's simple architecture, loss function, and training strategy. Furthermore, RecTable achieves high performance with fewer updates than other state-of-the-art methods. For example, TabDiff, one of the leading approaches, is trained for 8000 epochs. By contrast, RecTable is trained for 30k iterations, which means approximately 4313 epochs with a batch size of 4096. This suggests that RecTable can achieve high performance even with a relatively small number of updates.

\begin{table}[!t] 
   \centering
   \caption{Training time comparison. We use the implementation of the baselines provided by \citet{zhang2024mixedtype} and \citet{shi2025tabdiff}.}
   \label{table:exp-time}
   \resizebox{\columnwidth}{!}{
   \begin{tabular}{lcccccccc}
      \toprule[0.8pt]
      Methods & GOGGLE & STaSy & GReaT & CoDi & TabDDPM & TabSyn & TabDiff & RecTable \\
      \midrule
      Training Time (sec) & 12752 & 6445 & 13773 & 4945 & 25910 & 2393 & 6231 & \textbf{1800} \\
   \bottomrule[1.0pt] 
   \end{tabular}
   }
\end{table}

\section{Ablation Studies}
We use the adult dataset for ablation studies.

\subsection{Backbone Architecture}
We compare four backbone architectures; TabDDPM (MLP-based), STaSy (ConcatSquashLinear-based), TabDiff (Transfomrers and MLP-based), and RecTable. As shown in \Cref{table:exp-abl-arc}, TabDDPM is competitive results with RecTable. However, as demonstrated in \Cref{table:exp-mle}, RecTable outperforms TabDDPM on the news dataset, suggests that the GLU-based architecture is more effective than the MLP-based. STaSy, whose architecture is inspired by \citet{grathwohl2018scalable}, does not achieve high performance. TabDiff, consisting of Transfomrers and MLPs, requires long training time due to the high computational cost of the attention mechanism. Moreover, despite the long training time, its AUC score is not best of them. 

\begin{table}[!t] 
   \centering
   \caption{Results of MLE with different backbone architectures. Other settings are same as RecTable's.}
   \label{table:exp-abl-arc}
   \begin{tabular}{lcccc}
      \toprule[0.8pt]
      Architecture & TabDDPM & STaSy & TabDiff & RecTable \\
      \midrule
      AUC & .904{\tiny$\pm$.002} & .487{\tiny$\pm$.046} & .905{\tiny$\pm$.001} & .906{\tiny$\pm$.002} \\
      Training Time (sec) & 840 & 1425 & 2400 & 1800 \\
   \bottomrule[1.0pt] 
   \end{tabular}
\end{table}

\subsection{Detailed Settings}
We compare the detailed settings in RecTable. We conduct two configurations: (config A) using a gaussian distribution as the noise distribution $\pi_0$ and a logit-normal timestep distribution, and (config B) using a combination of a gaussian distribution and a uniform distribution as $\pi_0$ along with a uniform timestep distribution. The results are shown in \Cref{table:exp-abl-detailed}. The combination of a gaussian distribution and a uniform distribution is effective in improving the quality of generated samples. Although the application of a logit-normal timestep distribution has a slight effect on the quality of the generated data, its impact is not as significant as observed in Stable Diffusion 3~\cite{pmlr-v235-esser24a}. In addition, sampling timestep increases training time, which may be unnecessary if faster training is desired.

\begin{table}[!t] 
   \centering
   \caption{Results of MLE with different detailed settings.}
   \label{table:exp-abl-detailed}
   \begin{threeparttable}{
   \begin{tabular}{lccc}
      \toprule[0.8pt]
      Configuration & config A$^{3}$ & config B$^{4}$  & RecTable \\
      \midrule
      AUC & .890{\tiny$\pm$.005} & .905{\tiny$\pm$.001} & .906{\tiny$\pm$.002} \\
      Training Time (sec) & 409 & 600 & 1800 \\
   \bottomrule[1.0pt] 
   \end{tabular}
   }
   \begin{tablenotes}
      \item[3] RecTable without a combined distribution.
      \item[4] RecTable without a logit-normal timestep distribution.
   \end{tablenotes}
   \vspace{-5pt}
   \end{threeparttable}
\end{table}

\section{Conclusion}
In this paper, we presented RecTable for tabular data synthesis. RecTable incorporates rectified flow without reflow, simpler training strategies compared to diffusion models, a combination of a gaussian distribution for numerical data and a uniform distribution for categorical data, a gated linear unit-based architecture for fast training and high-quality generation, and a logit-normal timestep distribution. We evaluated RecTable against several state-of-the-art baselines on six real-world datasets and confirmed that it maintains or outperforms these baselines in machine learning efficiency and diverse generation. We believe that the rectified flow framework has potential for high-quality tabular data synthesis. Despite RecTable's simple architecture and training strategies, it achieves competitive results with state-of-the-art methods. This indicates that with more sophisticated architecture and training strategies, rectified flow could surpass denoising diffusion models.

\bibliography{iclr2025_conference}

\begin{thebibliography}{61}
\providecommand{\natexlab}[1]{#1}
\providecommand{\url}[1]{\texttt{#1}}
\expandafter\ifx\csname urlstyle\endcsname\relax
  \providecommand{\doi}[1]{doi: #1}\else
  \providecommand{\doi}{doi: \begingroup \urlstyle{rm}\Url}\fi

\bibitem[Alaa et~al.(2022)Alaa, Van~Breugel, Saveliev, and van~der
  Schaar]{pmlr-v162-alaa22a}
Ahmed Alaa, Boris Van~Breugel, Evgeny~S. Saveliev, and Mihaela van~der Schaar.
\newblock How faithful is your synthetic data? {S}ample-level metrics for
  evaluating and auditing generative models.
\newblock In Kamalika Chaudhuri, Stefanie Jegelka, Le~Song, Csaba Szepesvari,
  Gang Niu, and Sivan Sabato (eds.), \emph{Proceedings of the 39th
  International Conference on Machine Learning}, volume 162 of
  \emph{Proceedings of Machine Learning Research}, pp.\  290--306. PMLR, 17--23
  Jul 2022.
\newblock URL \url{https://proceedings.mlr.press/v162/alaa22a.html}.

\bibitem[Albergo \& Vanden-Eijnden(2023)Albergo and
  Vanden-Eijnden]{albergo2023building}
Michael~Samuel Albergo and Eric Vanden-Eijnden.
\newblock Building normalizing flows with stochastic interpolants.
\newblock In \emph{The Eleventh International Conference on Learning
  Representations}, 2023.
\newblock URL \url{https://openreview.net/forum?id=li7qeBbCR1t}.

\bibitem[ATCHISON \& SHEN(1980)ATCHISON and SHEN]{10.1093/biomet/67.2.261}
J.~ATCHISON and S.M. SHEN.
\newblock Logistic-normal distributions:some properties and uses.
\newblock \emph{Biometrika}, 67\penalty0 (2):\penalty0 261--272, 08 1980.
\newblock ISSN 0006-3444.
\newblock \doi{10.1093/biomet/67.2.261}.
\newblock URL \url{https://doi.org/10.1093/biomet/67.2.261}.

\bibitem[Becker \& Kohavi(1996)Becker and Kohavi]{adult_2}
Barry Becker and Ronny Kohavi.
\newblock {Adult}.
\newblock UCI Machine Learning Repository, 1996.
\newblock {DOI}: https://doi.org/10.24432/C5XW20.

\bibitem[Bock(2004)]{magic_gamma_telescope_159}
R.~Bock.
\newblock {MAGIC Gamma Telescope}.
\newblock UCI Machine Learning Repository, 2004.
\newblock {DOI}: https://doi.org/10.24432/C52C8B.

\bibitem[Bommasani et~al.(2022)Bommasani, Hudson, Adeli, Altman, Arora, von
  Arx, Bernstein, Bohg, Bosselut, Brunskill, Brynjolfsson, Buch, Card,
  Castellon, Chatterji, Chen, Creel, Davis, Demszky, Donahue, Doumbouya,
  Durmus, Ermon, Etchemendy, Ethayarajh, Fei-Fei, Finn, Gale, Gillespie, Goel,
  Goodman, Grossman, Guha, Hashimoto, Henderson, Hewitt, Ho, Hong, Hsu, Huang,
  Icard, Jain, Jurafsky, Kalluri, Karamcheti, Keeling, Khani, Khattab, Koh,
  Krass, Krishna, Kuditipudi, Kumar, Ladhak, Lee, Lee, Leskovec, Levent, Li,
  Li, Ma, Malik, Manning, Mirchandani, Mitchell, Munyikwa, Nair, Narayan,
  Narayanan, Newman, Nie, Niebles, Nilforoshan, Nyarko, Ogut, Orr,
  Papadimitriou, Park, Piech, Portelance, Potts, Raghunathan, Reich, Ren, Rong,
  Roohani, Ruiz, Ryan, Ré, Sadigh, Sagawa, Santhanam, Shih, Srinivasan,
  Tamkin, Taori, Thomas, Tramèr, Wang, Wang, Wu, Wu, Wu, Xie, Yasunaga, You,
  Zaharia, Zhang, Zhang, Zhang, Zhang, Zheng, Zhou, and
  Liang]{bommasani2022opportunitiesrisksfoundationmodels}
Rishi Bommasani, Drew~A. Hudson, Ehsan Adeli, Russ Altman, Simran Arora, Sydney
  von Arx, Michael~S. Bernstein, Jeannette Bohg, Antoine Bosselut, Emma
  Brunskill, Erik Brynjolfsson, Shyamal Buch, Dallas Card, Rodrigo Castellon,
  Niladri Chatterji, Annie Chen, Kathleen Creel, Jared~Quincy Davis, Dora
  Demszky, Chris Donahue, Moussa Doumbouya, Esin Durmus, Stefano Ermon, John
  Etchemendy, Kawin Ethayarajh, Li~Fei-Fei, Chelsea Finn, Trevor Gale, Lauren
  Gillespie, Karan Goel, Noah Goodman, Shelby Grossman, Neel Guha, Tatsunori
  Hashimoto, Peter Henderson, John Hewitt, Daniel~E. Ho, Jenny Hong, Kyle Hsu,
  Jing Huang, Thomas Icard, Saahil Jain, Dan Jurafsky, Pratyusha Kalluri,
  Siddharth Karamcheti, Geoff Keeling, Fereshte Khani, Omar Khattab, Pang~Wei
  Koh, Mark Krass, Ranjay Krishna, Rohith Kuditipudi, Ananya Kumar, Faisal
  Ladhak, Mina Lee, Tony Lee, Jure Leskovec, Isabelle Levent, Xiang~Lisa Li,
  Xuechen Li, Tengyu Ma, Ali Malik, Christopher~D. Manning, Suvir Mirchandani,
  Eric Mitchell, Zanele Munyikwa, Suraj Nair, Avanika Narayan, Deepak
  Narayanan, Ben Newman, Allen Nie, Juan~Carlos Niebles, Hamed Nilforoshan,
  Julian Nyarko, Giray Ogut, Laurel Orr, Isabel Papadimitriou, Joon~Sung Park,
  Chris Piech, Eva Portelance, Christopher Potts, Aditi Raghunathan, Rob Reich,
  Hongyu Ren, Frieda Rong, Yusuf Roohani, Camilo Ruiz, Jack Ryan, Christopher
  Ré, Dorsa Sadigh, Shiori Sagawa, Keshav Santhanam, Andy Shih, Krishnan
  Srinivasan, Alex Tamkin, Rohan Taori, Armin~W. Thomas, Florian Tramèr,
  Rose~E. Wang, William Wang, Bohan Wu, Jiajun Wu, Yuhuai Wu, Sang~Michael Xie,
  Michihiro Yasunaga, Jiaxuan You, Matei Zaharia, Michael Zhang, Tianyi Zhang,
  Xikun Zhang, Yuhui Zhang, Lucia Zheng, Kaitlyn Zhou, and Percy Liang.
\newblock On the opportunities and risks of foundation models, 2022.
\newblock URL \url{https://arxiv.org/abs/2108.07258}.

\bibitem[Borisov et~al.(2023)Borisov, Sessler, Leemann, Pawelczyk, and
  Kasneci]{borisov2023language}
Vadim Borisov, Kathrin Sessler, Tobias Leemann, Martin Pawelczyk, and Gjergji
  Kasneci.
\newblock Language models are realistic tabular data generators.
\newblock In \emph{The Eleventh International Conference on Learning
  Representations}, 2023.
\newblock URL \url{https://openreview.net/forum?id=cEygmQNOeI}.

\bibitem[Bunkhumpornpat et~al.(2009)Bunkhumpornpat, Sinapiromsaran, and
  Lursinsap]{10.1007/978-3-642-01307-2_43}
Chumphol Bunkhumpornpat, Krung Sinapiromsaran, and Chidchanok Lursinsap.
\newblock Safe-level-smote: Safe-level-synthetic minority over-sampling
  technique for handling the class imbalanced problem.
\newblock In Thanaruk Theeramunkong, Boonserm Kijsirikul, Nick Cercone, and
  Tu-Bao Ho (eds.), \emph{Advances in Knowledge Discovery and Data Mining},
  pp.\  475--482, Berlin, Heidelberg, 2009. Springer Berlin Heidelberg.
\newblock ISBN 978-3-642-01307-2.

\bibitem[Chawla et~al.(2002)Chawla, Bowyer, Hall, and
  Kegelmeyer]{10.5555/1622407.1622416}
Nitesh~V. Chawla, Kevin~W. Bowyer, Lawrence~O. Hall, and W.~Philip Kegelmeyer.
\newblock Smote: synthetic minority over-sampling technique.
\newblock \emph{J. Artif. Int. Res.}, 16\penalty0 (1):\penalty0 321--357, June
  2002.
\newblock ISSN 1076-9757.

\bibitem[Chen et~al.(2018)Chen, Rubanova, Bettencourt, and
  Duvenaud]{NEURIPS2018_69386f6b}
Ricky T.~Q. Chen, Yulia Rubanova, Jesse Bettencourt, and David~K Duvenaud.
\newblock Neural ordinary differential equations.
\newblock In S.~Bengio, H.~Wallach, H.~Larochelle, K.~Grauman, N.~Cesa-Bianchi,
  and R.~Garnett (eds.), \emph{Advances in Neural Information Processing
  Systems}, volume~31. Curran Associates, Inc., 2018.
\newblock URL
  \url{https://proceedings.neurips.cc/paper_files/paper/2018/file/69386f6bb1dfed68692a24c8686939b9-Paper.pdf}.

\bibitem[Chen(2015)]{beijing_pm2.5_381}
Song Chen.
\newblock {Beijing PM2.5}.
\newblock UCI Machine Learning Repository, 2015.
\newblock {DOI}: https://doi.org/10.24432/C5JS49.

\bibitem[Chen \& Guestrin(2016)Chen and Guestrin]{10.1145/2939672.2939785}
Tianqi Chen and Carlos Guestrin.
\newblock Xgboost: A scalable tree boosting system.
\newblock In \emph{Proceedings of the 22nd ACM SIGKDD International Conference
  on Knowledge Discovery and Data Mining}, KDD '16, pp.\  785--794, New York,
  NY, USA, 2016. Association for Computing Machinery.
\newblock ISBN 9781450342322.
\newblock \doi{10.1145/2939672.2939785}.
\newblock URL \url{https://doi.org/10.1145/2939672.2939785}.

\bibitem[Chow \& Liu(1968)Chow and Liu]{1054142}
C.~Chow and C.~Liu.
\newblock Approximating discrete probability distributions with dependence
  trees.
\newblock \emph{IEEE Transactions on Information Theory}, 14\penalty0
  (3):\penalty0 462--467, 1968.
\newblock \doi{10.1109/TIT.1968.1054142}.

\bibitem[Dat(2023)]{sdmetrics}
\emph{Synthetic Data Metrics}.
\newblock DataCebo, Inc., 2023.

\bibitem[Dauphin et~al.(2017)Dauphin, Fan, Auli, and
  Grangier]{pmlr-v70-dauphin17a}
Yann~N. Dauphin, Angela Fan, Michael Auli, and David Grangier.
\newblock Language modeling with gated convolutional networks.
\newblock In Doina Precup and Yee~Whye Teh (eds.), \emph{Proceedings of the
  34th International Conference on Machine Learning}, volume~70 of
  \emph{Proceedings of Machine Learning Research}, pp.\  933--941. PMLR, 06--11
  Aug 2017.
\newblock URL \url{https://proceedings.mlr.press/v70/dauphin17a.html}.

\bibitem[Dhariwal \& Nichol(2021)Dhariwal and Nichol]{dhariwal2021diffusion}
Prafulla Dhariwal and Alexander~Quinn Nichol.
\newblock Diffusion models beat {GAN}s on image synthesis.
\newblock In A.~Beygelzimer, Y.~Dauphin, P.~Liang, and J.~Wortman Vaughan
  (eds.), \emph{Advances in Neural Information Processing Systems}, 2021.
\newblock URL \url{https://openreview.net/forum?id=AAWuCvzaVt}.

\bibitem[Esser et~al.(2024)Esser, Kulal, Blattmann, Entezari, M\"{u}ller,
  Saini, Levi, Lorenz, Sauer, Boesel, Podell, Dockhorn, English, and
  Rombach]{pmlr-v235-esser24a}
Patrick Esser, Sumith Kulal, Andreas Blattmann, Rahim Entezari, Jonas
  M\"{u}ller, Harry Saini, Yam Levi, Dominik Lorenz, Axel Sauer, Frederic
  Boesel, Dustin Podell, Tim Dockhorn, Zion English, and Robin Rombach.
\newblock Scaling rectified flow transformers for high-resolution image
  synthesis.
\newblock In Ruslan Salakhutdinov, Zico Kolter, Katherine Heller, Adrian
  Weller, Nuria Oliver, Jonathan Scarlett, and Felix Berkenkamp (eds.),
  \emph{Proceedings of the 41st International Conference on Machine Learning},
  volume 235 of \emph{Proceedings of Machine Learning Research}, pp.\
  12606--12633. PMLR, 21--27 Jul 2024.
\newblock URL \url{https://proceedings.mlr.press/v235/esser24a.html}.

\bibitem[Fernandes \& Sernadela(2015)Fernandes and
  Sernadela]{online_news_popularity_332}
Vinagre Pedro Cortez~Paulo Fernandes, Kelwin and Pedro Sernadela.
\newblock {Online News Popularity}.
\newblock UCI Machine Learning Repository, 2015.
\newblock {DOI}: https://doi.org/10.24432/C5NS3V.

\bibitem[Fonseca \& Bacao(2023)Fonseca and Bacao]{Fonseca2023}
Joao Fonseca and Fernando Bacao.
\newblock Tabular and latent space synthetic data generation: a literature
  review.
\newblock \emph{Journal of Big Data}, 10\penalty0 (1):\penalty0 115, Jul 2023.
\newblock ISSN 2196-1115.
\newblock \doi{10.1186/s40537-023-00792-7}.
\newblock URL \url{https://doi.org/10.1186/s40537-023-00792-7}.

\bibitem[Ghasemi \& Amyot(2016)Ghasemi and Amyot]{doi:10.1504/IJEH.2016.078745}
Mahdi Ghasemi and Daniel Amyot.
\newblock Process mining in healthcare: a systematised literature review.
\newblock \emph{International Journal of Electronic Healthcare}, 9\penalty0
  (1):\penalty0 60--88, 2016.
\newblock \doi{10.1504/IJEH.2016.078745}.
\newblock URL
  \url{https://www.inderscienceonline.com/doi/abs/10.1504/IJEH.2016.078745}.

\bibitem[Goodfellow et~al.(2014)Goodfellow, Pouget-Abadie, Mirza, Xu,
  Warde-Farley, Ozair, Courville, and Bengio]{NIPS2014_f033ed80}
Ian~J. Goodfellow, Jean Pouget-Abadie, Mehdi Mirza, Bing Xu, David
  Warde-Farley, Sherjil Ozair, Aaron Courville, and Yoshua Bengio.
\newblock Generative adversarial nets.
\newblock In Z.~Ghahramani, M.~Welling, C.~Cortes, N.~Lawrence, and K.Q.
  Weinberger (eds.), \emph{Advances in Neural Information Processing Systems},
  volume~27. Curran Associates, Inc., 2014.
\newblock URL
  \url{https://proceedings.neurips.cc/paper_files/paper/2014/file/f033ed80deb0234979a61f95710dbe25-Paper.pdf}.

\bibitem[Grathwohl et~al.(2019)Grathwohl, Chen, Bettencourt, and
  Duvenaud]{grathwohl2018scalable}
Will Grathwohl, Ricky T.~Q. Chen, Jesse Bettencourt, and David Duvenaud.
\newblock Scalable reversible generative models with free-form continuous
  dynamics.
\newblock In \emph{International Conference on Learning Representations}, 2019.
\newblock URL \url{https://openreview.net/forum?id=rJxgknCcK7}.

\bibitem[Guan et~al.(2024)Guan, Su, Zhou, Miao, Xie, Li, and Hong]{10447822}
Wenhao Guan, Qi~Su, Haodong Zhou, Shiyu Miao, Xingjia Xie, Lin Li, and Qingyang
  Hong.
\newblock Reflow-tts: A rectified flow model for high-fidelity text-to-speech.
\newblock In \emph{ICASSP 2024 - 2024 IEEE International Conference on
  Acoustics, Speech and Signal Processing (ICASSP)}, pp.\  10501--10505, 2024.
\newblock \doi{10.1109/ICASSP48485.2024.10447822}.

\bibitem[Guo et~al.(2024)Guo, Du, Ma, Chen, and Yu]{10445948}
Yiwei Guo, Chenpeng Du, Ziyang Ma, Xie Chen, and Kai Yu.
\newblock Voiceflow: Efficient text-to-speech with rectified flow matching.
\newblock In \emph{ICASSP 2024 - 2024 IEEE International Conference on
  Acoustics, Speech and Signal Processing (ICASSP)}, pp.\  11121--11125, 2024.
\newblock \doi{10.1109/ICASSP48485.2024.10445948}.

\bibitem[Han et~al.(2005)Han, Wang, and Mao]{10.1007/11538059_91}
Hui Han, Wen-Yuan Wang, and Bing-Huan Mao.
\newblock Borderline-smote: A new over-sampling method in imbalanced data sets
  learning.
\newblock In De-Shuang Huang, Xiao-Ping Zhang, and Guang-Bin Huang (eds.),
  \emph{Advances in Intelligent Computing}, pp.\  878--887, Berlin, Heidelberg,
  2005. Springer Berlin Heidelberg.
\newblock ISBN 978-3-540-31902-3.

\bibitem[Ho et~al.(2020)Ho, Jain, and Abbeel]{NEURIPS2020_4c5bcfec}
Jonathan Ho, Ajay Jain, and Pieter Abbeel.
\newblock Denoising diffusion probabilistic models.
\newblock In H.~Larochelle, M.~Ranzato, R.~Hadsell, M.F. Balcan, and H.~Lin
  (eds.), \emph{Advances in Neural Information Processing Systems}, volume~33,
  pp.\  6840--6851. Curran Associates, Inc., 2020.
\newblock URL
  \url{https://proceedings.neurips.cc/paper_files/paper/2020/file/4c5bcfec8584af0d967f1ab10179ca4b-Paper.pdf}.

\bibitem[Kim et~al.(2023)Kim, Lee, and Park]{kim2023stasy}
Jayoung Kim, Chaejeong Lee, and Noseong Park.
\newblock {ST}asy: Score-based tabular data synthesis.
\newblock In \emph{The Eleventh International Conference on Learning
  Representations}, 2023.
\newblock URL \url{https://openreview.net/forum?id=1mNssCWt_v}.

\bibitem[Kingma \& Ba(2017)Kingma and
  Ba]{kingma2017adammethodstochasticoptimization}
Diederik~P. Kingma and Jimmy Ba.
\newblock Adam: A method for stochastic optimization, 2017.
\newblock URL \url{https://arxiv.org/abs/1412.6980}.

\bibitem[Kingma \& Welling(2022)Kingma and
  Welling]{kingma2022autoencodingvariationalbayes}
Diederik~P Kingma and Max Welling.
\newblock Auto-encoding variational bayes, 2022.
\newblock URL \url{https://arxiv.org/abs/1312.6114}.

\bibitem[Kotelnikov et~al.(2023)Kotelnikov, Baranchuk, Rubachev, and
  Babenko]{pmlr-v202-kotelnikov23a}
Akim Kotelnikov, Dmitry Baranchuk, Ivan Rubachev, and Artem Babenko.
\newblock {T}ab{DDPM}: Modelling tabular data with diffusion models.
\newblock In Andreas Krause, Emma Brunskill, Kyunghyun Cho, Barbara Engelhardt,
  Sivan Sabato, and Jonathan Scarlett (eds.), \emph{Proceedings of the 40th
  International Conference on Machine Learning}, volume 202 of
  \emph{Proceedings of Machine Learning Research}, pp.\  17564--17579. PMLR,
  23--29 Jul 2023.
\newblock URL \url{https://proceedings.mlr.press/v202/kotelnikov23a.html}.

\bibitem[Lee et~al.(2023)Lee, Kim, and Park]{pmlr-v202-lee23i}
Chaejeong Lee, Jayoung Kim, and Noseong Park.
\newblock {C}o{D}i: Co-evolving contrastive diffusion models for mixed-type
  tabular synthesis.
\newblock In Andreas Krause, Emma Brunskill, Kyunghyun Cho, Barbara Engelhardt,
  Sivan Sabato, and Jonathan Scarlett (eds.), \emph{Proceedings of the 40th
  International Conference on Machine Learning}, volume 202 of
  \emph{Proceedings of Machine Learning Research}, pp.\  18940--18956. PMLR,
  23--29 Jul 2023.
\newblock URL \url{https://proceedings.mlr.press/v202/lee23i.html}.

\bibitem[Li et~al.(2021)Li, Huang, Yan, Zhou, Ye, and
  Liu]{10.1007/978-3-030-68790-8_50}
Yiren Li, Zheng Huang, Junchi Yan, Yi~Zhou, Fan Ye, and Xianhui Liu.
\newblock Gfte: Graph-based financial table extraction.
\newblock In Alberto Del~Bimbo, Rita Cucchiara, Stan Sclaroff, Giovanni~Maria
  Farinella, Tao Mei, Marco Bertini, Hugo~Jair Escalante, and Roberto Vezzani
  (eds.), \emph{Pattern Recognition. ICPR International Workshops and
  Challenges}, pp.\  644--658, Cham, 2021. Springer International Publishing.
\newblock ISBN 978-3-030-68790-8.

\bibitem[Lipman et~al.(2023)Lipman, Chen, Ben-Hamu, Nickel, and
  Le]{lipman2023flow}
Yaron Lipman, Ricky T.~Q. Chen, Heli Ben-Hamu, Maximilian Nickel, and Matthew
  Le.
\newblock Flow matching for generative modeling.
\newblock In \emph{The Eleventh International Conference on Learning
  Representations}, 2023.
\newblock URL \url{https://openreview.net/forum?id=PqvMRDCJT9t}.

\bibitem[Liu et~al.(2023{\natexlab{a}})Liu, Qian, Berrevoets, and van~der
  Schaar]{liu2023goggle}
Tennison Liu, Zhaozhi Qian, Jeroen Berrevoets, and Mihaela van~der Schaar.
\newblock {GOGGLE}: Generative modelling for tabular data by learning
  relational structure.
\newblock In \emph{The Eleventh International Conference on Learning
  Representations}, 2023{\natexlab{a}}.
\newblock URL \url{https://openreview.net/forum?id=fPVRcJqspu}.

\bibitem[Liu et~al.(2023{\natexlab{b}})Liu, Gong, and qiang liu]{liu2023flow}
Xingchao Liu, Chengyue Gong, and qiang liu.
\newblock Flow straight and fast: Learning to generate and transfer data with
  rectified flow.
\newblock In \emph{The Eleventh International Conference on Learning
  Representations}, 2023{\natexlab{b}}.
\newblock URL \url{https://openreview.net/forum?id=XVjTT1nw5z}.

\bibitem[Liu et~al.(2024)Liu, Zhang, Ma, Peng, and qiang liu]{liu2024instaflow}
Xingchao Liu, Xiwen Zhang, Jianzhu Ma, Jian Peng, and qiang liu.
\newblock Instaflow: One step is enough for high-quality diffusion-based
  text-to-image generation.
\newblock In \emph{The Twelfth International Conference on Learning
  Representations}, 2024.
\newblock URL \url{https://openreview.net/forum?id=1k4yZbbDqX}.

\bibitem[Lopez-Paz \& Oquab(2017)Lopez-Paz and Oquab]{lopez-paz2017revisiting}
David Lopez-Paz and Maxime Oquab.
\newblock Revisiting classifier two-sample tests.
\newblock In \emph{International Conference on Learning Representations}, 2017.
\newblock URL \url{https://openreview.net/forum?id=SJkXfE5xx}.

\bibitem[Mueller et~al.(2025)Mueller, Gruber, and Fok]{mueller2025continuous}
Markus Mueller, Kathrin Gruber, and Dennis Fok.
\newblock Continuous diffusion for mixed-type tabular data.
\newblock In \emph{The Thirteenth International Conference on Learning
  Representations}, 2025.
\newblock URL \url{https://openreview.net/forum?id=QPtoBPn4lZ}.

\bibitem[Mukherjee \& Khushi(2021)Mukherjee and Khushi]{Mukherjee_2021}
Mimi Mukherjee and Matloob Khushi.
\newblock Smote-enc: A novel smote-based method to generate synthetic data for
  nominal and continuous features.
\newblock \emph{Applied System Innovation}, 4\penalty0 (1):\penalty0 18, March
  2021.
\newblock ISSN 2571-5577.
\newblock \doi{10.3390/asi4010018}.
\newblock URL \url{http://dx.doi.org/10.3390/asi4010018}.

\bibitem[Narang et~al.(2021)Narang, Chung, Tay, Fedus, Fevry, Matena, Malkan,
  Fiedel, Shazeer, Lan, Zhou, Li, Ding, Marcus, Roberts, and
  Raffel]{narang-etal-2021-transformer}
Sharan Narang, Hyung~Won Chung, Yi~Tay, Liam Fedus, Thibault Fevry, Michael
  Matena, Karishma Malkan, Noah Fiedel, Noam Shazeer, Zhenzhong Lan, Yanqi
  Zhou, Wei Li, Nan Ding, Jake Marcus, Adam Roberts, and Colin Raffel.
\newblock Do transformer modifications transfer across implementations and
  applications?
\newblock In Marie-Francine Moens, Xuanjing Huang, Lucia Specia, and Scott
  Wen-tau Yih (eds.), \emph{Proceedings of the 2021 Conference on Empirical
  Methods in Natural Language Processing}, pp.\  5758--5773, Online and Punta
  Cana, Dominican Republic, November 2021. Association for Computational
  Linguistics.
\newblock \doi{10.18653/v1/2021.emnlp-main.465}.
\newblock URL \url{https://aclanthology.org/2021.emnlp-main.465/}.

\bibitem[Nederstigt et~al.(2014)Nederstigt, Aanen, Vandic, and
  Frasincar]{NEDERSTIGT2014296}
Lennart~J. Nederstigt, Steven~S. Aanen, Damir Vandic, and Flavius Frasincar.
\newblock Floppies: A framework for large-scale ontology population of product
  information from tabular data in e-commerce stores.
\newblock \emph{Decision Support Systems}, 59:\penalty0 296--311, 2014.
\newblock ISSN 0167-9236.
\newblock \doi{https://doi.org/10.1016/j.dss.2014.01.001}.
\newblock URL
  \url{https://www.sciencedirect.com/science/article/pii/S0167923614000025}.

\bibitem[Paszke et~al.(2019)Paszke, Gross, Massa, Lerer, Bradbury, Chanan,
  Killeen, Lin, Gimelshein, Antiga, Desmaison, Kopf, Yang, DeVito, Raison,
  Tejani, Chilamkurthy, Steiner, Fang, Bai, and Chintala]{NEURIPS2019_bdbca288}
Adam Paszke, Sam Gross, Francisco Massa, Adam Lerer, James Bradbury, Gregory
  Chanan, Trevor Killeen, Zeming Lin, Natalia Gimelshein, Luca Antiga, Alban
  Desmaison, Andreas Kopf, Edward Yang, Zachary DeVito, Martin Raison, Alykhan
  Tejani, Sasank Chilamkurthy, Benoit Steiner, Lu~Fang, Junjie Bai, and Soumith
  Chintala.
\newblock Pytorch: An imperative style, high-performance deep learning library.
\newblock In H.~Wallach, H.~Larochelle, A.~Beygelzimer, F.~d\textquotesingle
  Alch\'{e}-Buc, E.~Fox, and R.~Garnett (eds.), \emph{Advances in Neural
  Information Processing Systems}, volume~32. Curran Associates, Inc., 2019.
\newblock URL
  \url{https://proceedings.neurips.cc/paper_files/paper/2019/file/bdbca288fee7f92f2bfa9f7012727740-Paper.pdf}.

\bibitem[Pedregosa et~al.(2011)Pedregosa, Varoquaux, Gramfort, Michel, Thirion,
  Grisel, Blondel, Prettenhofer, Weiss, Dubourg, Vanderplas, Passos,
  Cournapeau, Brucher, Perrot, and {{\'E}}douard
  Duchesnay]{JMLR:v12:pedregosa11a}
Fabian Pedregosa, Ga{{\"e}}l Varoquaux, Alexandre Gramfort, Vincent Michel,
  Bertrand Thirion, Olivier Grisel, Mathieu Blondel, Peter Prettenhofer, Ron
  Weiss, Vincent Dubourg, Jake Vanderplas, Alexandre Passos, David Cournapeau,
  Matthieu Brucher, Matthieu Perrot, and {{\'E}}douard Duchesnay.
\newblock Scikit-learn: Machine learning in python.
\newblock \emph{Journal of Machine Learning Research}, 12\penalty0
  (85):\penalty0 2825--2830, 2011.
\newblock URL \url{http://jmlr.org/papers/v12/pedregosa11a.html}.

\bibitem[Radford et~al.(2019)Radford, Wu, Child, Luan, Amodei, and
  Sutskever]{radford2019language}
Alec Radford, Jeff Wu, Rewon Child, David Luan, Dario Amodei, and Ilya
  Sutskever.
\newblock Language models are unsupervised multitask learners.
\newblock 2019.

\bibitem[Rombach et~al.(2022)Rombach, Blattmann, Lorenz, Esser, and
  Ommer]{Rombach_2022_CVPR}
Robin Rombach, Andreas Blattmann, Dominik Lorenz, Patrick Esser, and Bj\"orn
  Ommer.
\newblock High-resolution image synthesis with latent diffusion models.
\newblock In \emph{Proceedings of the IEEE/CVF Conference on Computer Vision
  and Pattern Recognition (CVPR)}, pp.\  10684--10695, June 2022.

\bibitem[Saharia et~al.(2022)Saharia, Chan, Saxena, Li, Whang, Denton,
  Ghasemipour, Gontijo-Lopes, Ayan, Salimans, Ho, Fleet, and
  Norouzi]{saharia2022photorealistic}
Chitwan Saharia, William Chan, Saurabh Saxena, Lala Li, Jay Whang, Emily
  Denton, Seyed Kamyar~Seyed Ghasemipour, Raphael Gontijo-Lopes, Burcu~Karagol
  Ayan, Tim Salimans, Jonathan Ho, David~J. Fleet, and Mohammad Norouzi.
\newblock Photorealistic text-to-image diffusion models with deep language
  understanding.
\newblock In Alice~H. Oh, Alekh Agarwal, Danielle Belgrave, and Kyunghyun Cho
  (eds.), \emph{Advances in Neural Information Processing Systems}, 2022.
\newblock URL \url{https://openreview.net/forum?id=08Yk-n5l2Al}.

\bibitem[Sakar \& Kastro(2018)Sakar and
  Kastro]{online_shoppers_purchasing_intention_dataset_468}
C.~Sakar and Yomi Kastro.
\newblock {Online Shoppers Purchasing Intention Dataset}.
\newblock UCI Machine Learning Repository, 2018.
\newblock {DOI}: https://doi.org/10.24432/C5F88Q.

\bibitem[Shenkar \& Wolf(2022)Shenkar and Wolf]{shenkar2022anomaly}
Tom Shenkar and Lior Wolf.
\newblock Anomaly detection for tabular data with internal contrastive
  learning.
\newblock In \emph{International Conference on Learning Representations}, 2022.
\newblock URL \url{https://openreview.net/forum?id=_hszZbt46bT}.

\bibitem[Shi et~al.(2025)Shi, Xu, Hua, Zhang, Ermon, and
  Leskovec]{shi2025tabdiff}
Juntong Shi, Minkai Xu, Harper Hua, Hengrui Zhang, Stefano Ermon, and Jure
  Leskovec.
\newblock Tabdiff: a mixed-type diffusion model for tabular data generation.
\newblock In \emph{The Thirteenth International Conference on Learning
  Representations}, 2025.
\newblock URL \url{https://openreview.net/forum?id=swvURjrt8z}.

\bibitem[Sohl-Dickstein et~al.(2015)Sohl-Dickstein, Weiss, Maheswaranathan, and
  Ganguli]{pmlr-v37-sohl-dickstein15}
Jascha Sohl-Dickstein, Eric Weiss, Niru Maheswaranathan, and Surya Ganguli.
\newblock Deep unsupervised learning using nonequilibrium thermodynamics.
\newblock In Francis Bach and David Blei (eds.), \emph{Proceedings of the 32nd
  International Conference on Machine Learning}, volume~37 of \emph{Proceedings
  of Machine Learning Research}, pp.\  2256--2265, Lille, France, 07--09 Jul
  2015. PMLR.
\newblock URL \url{https://proceedings.mlr.press/v37/sohl-dickstein15.html}.

\bibitem[Srivastava et~al.(2017)Srivastava, Valkov, Russell, Gutmann, and
  Sutton]{NIPS2017_44a2e080}
Akash Srivastava, Lazar Valkov, Chris Russell, Michael~U. Gutmann, and Charles
  Sutton.
\newblock Veegan: Reducing mode collapse in gans using implicit variational
  learning.
\newblock In I.~Guyon, U.~Von Luxburg, S.~Bengio, H.~Wallach, R.~Fergus,
  S.~Vishwanathan, and R.~Garnett (eds.), \emph{Advances in Neural Information
  Processing Systems}, volume~30. Curran Associates, Inc., 2017.
\newblock URL
  \url{https://proceedings.neurips.cc/paper_files/paper/2017/file/44a2e0804995faf8d2e3b084a1e2db1d-Paper.pdf}.

\bibitem[Vaswani et~al.(2017)Vaswani, Shazeer, Parmar, Uszkoreit, Jones, Gomez,
  Kaiser, and Polosukhin]{NIPS2017_3f5ee243}
Ashish Vaswani, Noam Shazeer, Niki Parmar, Jakob Uszkoreit, Llion Jones,
  Aidan~N Gomez, \L~ukasz Kaiser, and Illia Polosukhin.
\newblock Attention is all you need.
\newblock In I.~Guyon, U.~Von Luxburg, S.~Bengio, H.~Wallach, R.~Fergus,
  S.~Vishwanathan, and R.~Garnett (eds.), \emph{Advances in Neural Information
  Processing Systems}, volume~30. Curran Associates, Inc., 2017.
\newblock URL
  \url{https://proceedings.neurips.cc/paper_files/paper/2017/file/3f5ee243547dee91fbd053c1c4a845aa-Paper.pdf}.

\bibitem[Virtanen et~al.(2020)Virtanen, Gommers, Oliphant, Haberland, Reddy,
  Cournapeau, Burovski, Peterson, Weckesser, Bright, van~der Walt, Brett,
  Wilson, Millman, Mayorov, Nelson, Jones, Kern, Larson, Carey, Polat, Feng,
  Moore, VanderPlas, Laxalde, Perktold, Cimrman, Henriksen, Quintero, Harris,
  Archibald, Ribeiro, Pedregosa, van Mulbregt, Vijaykumar, Bardelli, Rothberg,
  Hilboll, Kloeckner, Scopatz, Lee, Rokem, Woods, Fulton, Masson,
  H{\"a}ggstr{\"o}m, Fitzgerald, Nicholson, Hagen, Pasechnik, Olivetti, Martin,
  Wieser, Silva, Lenders, Wilhelm, Young, Price, Ingold, Allen, Lee, Audren,
  Probst, Dietrich, Silterra, Webber, Slavi{\v{c}}, Nothman, Buchner, Kulick,
  Sch{\"o}nberger, de~Miranda~Cardoso, Reimer, Harrington, Rodr{\'i}guez,
  Nunez-Iglesias, Kuczynski, Tritz, Thoma, Newville, K{\"u}mmerer, Bolingbroke,
  Tartre, Pak, Smith, Nowaczyk, Shebanov, Pavlyk, Brodtkorb, Lee, McGibbon,
  Feldbauer, Lewis, Tygier, Sievert, Vigna, Peterson, More, Pudlik, Oshima,
  Pingel, Robitaille, Spura, Jones, Cera, Leslie, Zito, Krauss, Upadhyay,
  Halchenko, V{\'a}zquez-Baeza, and 1.0~Contributors]{Virtanen2020}
Pauli Virtanen, Ralf Gommers, Travis~E. Oliphant, Matt Haberland, Tyler Reddy,
  David Cournapeau, Evgeni Burovski, Pearu Peterson, Warren Weckesser, Jonathan
  Bright, St{\'e}fan~J. van~der Walt, Matthew Brett, Joshua Wilson, K.~Jarrod
  Millman, Nikolay Mayorov, Andrew R.~J. Nelson, Eric Jones, Robert Kern, Eric
  Larson, C.~J. Carey, {\.{I}}lhan Polat, Yu~Feng, Eric~W. Moore, Jake
  VanderPlas, Denis Laxalde, Josef Perktold, Robert Cimrman, Ian Henriksen,
  E.~A. Quintero, Charles~R. Harris, Anne~M. Archibald, Ant{\^o}nio~H. Ribeiro,
  Fabian Pedregosa, Paul van Mulbregt, Aditya Vijaykumar, Alessandro~Pietro
  Bardelli, Alex Rothberg, Andreas Hilboll, Andreas Kloeckner, Anthony Scopatz,
  Antony Lee, Ariel Rokem, C.~Nathan Woods, Chad Fulton, Charles Masson,
  Christian H{\"a}ggstr{\"o}m, Clark Fitzgerald, David~A. Nicholson, David~R.
  Hagen, Dmitrii~V. Pasechnik, Emanuele Olivetti, Eric Martin, Eric Wieser,
  Fabrice Silva, Felix Lenders, Florian Wilhelm, G.~Young, Gavin~A. Price,
  Gert-Ludwig Ingold, Gregory~E. Allen, Gregory~R. Lee, Herv{\'e} Audren, Irvin
  Probst, J{\"o}rg~P. Dietrich, Jacob Silterra, James~T. Webber, Janko
  Slavi{\v{c}}, Joel Nothman, Johannes Buchner, Johannes Kulick, Johannes~L.
  Sch{\"o}nberger, Jos{\'e}~Vin{\'i}cius de~Miranda~Cardoso, Joscha Reimer,
  Joseph Harrington, Juan Luis~Cano Rodr{\'i}guez, Juan Nunez-Iglesias, Justin
  Kuczynski, Kevin Tritz, Martin Thoma, Matthew Newville, Matthias
  K{\"u}mmerer, Maximilian Bolingbroke, Michael Tartre, Mikhail Pak,
  Nathaniel~J. Smith, Nikolai Nowaczyk, Nikolay Shebanov, Oleksandr Pavlyk,
  Per~A. Brodtkorb, Perry Lee, Robert~T. McGibbon, Roman Feldbauer, Sam Lewis,
  Sam Tygier, Scott Sievert, Sebastiano Vigna, Stefan Peterson, Surhud More,
  Tadeusz Pudlik, Takuya Oshima, Thomas~J. Pingel, Thomas~P. Robitaille, Thomas
  Spura, Thouis~R. Jones, Tim Cera, Tim Leslie, Tiziano Zito, Tom Krauss,
  Utkarsh Upadhyay, Yaroslav~O. Halchenko, Yoshiki V{\'a}zquez-Baeza, and SciPy
  1.0~Contributors.
\newblock Scipy 1.0: fundamental algorithms for scientific computing in python.
\newblock \emph{Nature Methods}, 17\penalty0 (3):\penalty0 261--272, Mar 2020.
\newblock ISSN 1548-7105.
\newblock \doi{10.1038/s41592-019-0686-2}.
\newblock URL \url{https://doi.org/10.1038/s41592-019-0686-2}.

\bibitem[Wang et~al.(2025)Wang, Yang, Huang, Wang, and Li]{wang2025rectified}
Fu-Yun Wang, Ling Yang, Zhaoyang Huang, Mengdi Wang, and Hongsheng Li.
\newblock Rectified diffusion: Straightness is not your need in rectified flow.
\newblock In \emph{The Thirteenth International Conference on Learning
  Representations}, 2025.
\newblock URL \url{https://openreview.net/forum?id=nEDToD1R8M}.

\bibitem[Wang et~al.(2024{\natexlab{a}})Wang, Guo, Huang, Huang, Wang, You, Li,
  and Zhao]{wang2024frieren}
Yongqi Wang, Wenxiang Guo, Rongjie Huang, Jiawei Huang, Zehan Wang, Fuming You,
  Ruiqi Li, and Zhou Zhao.
\newblock Frieren: Efficient video-to-audio generation network with rectified
  flow matching.
\newblock In \emph{The Thirty-eighth Annual Conference on Neural Information
  Processing Systems}, 2024{\natexlab{a}}.
\newblock URL \url{https://openreview.net/forum?id=prXfM5X2Db}.

\bibitem[Wang et~al.(2024{\natexlab{b}})Wang, Gao, Xiao, and
  Sun]{10.24963/ijcai.2024/670}
Zifeng Wang, Chufan Gao, Cao Xiao, and Jimeng Sun.
\newblock Meditab: scaling medical tabular data predictors via data
  consolidation, enrichment, and refinement.
\newblock In \emph{Proceedings of the Thirty-Third International Joint
  Conference on Artificial Intelligence}, IJCAI '24, 2024{\natexlab{b}}.
\newblock ISBN 978-1-956792-04-1.
\newblock \doi{10.24963/ijcai.2024/670}.
\newblock URL \url{https://doi.org/10.24963/ijcai.2024/670}.

\bibitem[Xu et~al.(2019)Xu, Skoularidou, Cuesta-Infante, and
  Veeramachaneni]{NEURIPS2019_254ed7d2}
Lei Xu, Maria Skoularidou, Alfredo Cuesta-Infante, and Kalyan Veeramachaneni.
\newblock Modeling tabular data using conditional gan.
\newblock In H.~Wallach, H.~Larochelle, A.~Beygelzimer, F.~d\textquotesingle
  Alch\'{e}-Buc, E.~Fox, and R.~Garnett (eds.), \emph{Advances in Neural
  Information Processing Systems}, volume~32. Curran Associates, Inc., 2019.
\newblock URL
  \url{https://proceedings.neurips.cc/paper_files/paper/2019/file/254ed7d2de3b23ab10936522dd547b78-Paper.pdf}.

\bibitem[Yeh(2009)]{default_of_credit_card_clients_350}
I-Cheng Yeh.
\newblock {Default of Credit Card Clients}.
\newblock UCI Machine Learning Repository, 2009.
\newblock {DOI}: https://doi.org/10.24432/C55S3H.

\bibitem[Zhang et~al.(2024{\natexlab{a}})Zhang, Zhang, Shen, Srinivasan, Qin,
  Faloutsos, Rangwala, and Karypis]{zhang2024mixedtype}
Hengrui Zhang, Jiani Zhang, Zhengyuan Shen, Balasubramaniam Srinivasan, Xiao
  Qin, Christos Faloutsos, Huzefa Rangwala, and George Karypis.
\newblock Mixed-type tabular data synthesis with score-based diffusion in
  latent space.
\newblock In \emph{The Twelfth International Conference on Learning
  Representations}, 2024{\natexlab{a}}.
\newblock URL \url{https://openreview.net/forum?id=4Ay23yeuz0}.

\bibitem[Zhang et~al.(2024{\natexlab{b}})Zhang, Wu, Gong, and
  Liu]{zhang-etal-2024-languageflow}
Shujian Zhang, Lemeng Wu, Chengyue Gong, and Xingchao Liu.
\newblock {L}anguage{F}low: Advancing diffusion language generation with
  probabilistic flows.
\newblock In Kevin Duh, Helena Gomez, and Steven Bethard (eds.),
  \emph{Proceedings of the 2024 Conference of the North American Chapter of the
  Association for Computational Linguistics: Human Language Technologies
  (Volume 1: Long Papers)}, pp.\  3893--3905, Mexico City, Mexico, June
  2024{\natexlab{b}}. Association for Computational Linguistics.
\newblock \doi{10.18653/v1/2024.naacl-long.215}.
\newblock URL \url{https://aclanthology.org/2024.naacl-long.215/}.

\bibitem[Zheng \& Charoenphakdee(2022)Zheng and
  Charoenphakdee]{tashiro2021csdi}
Shuhan Zheng and Nontawat Charoenphakdee.
\newblock Diffusion models for missing value imputation in tabular data.
\newblock In \emph{NeurIPS Table Representation Learning (TRL) Workshop}, 2022.

\end{thebibliography}
\bibliographystyle{iclr2025_conference}

\clearpage
\appendix
\section{Datasets}
\label{appendix:datasets}
We use six real-world tabular datasets: Adult, Default, Shoppers, Magic, Faults, Beijing, and News from UCI Machine Learning Repository\footnote{\url{https://archive.ics.uci.edu/datasets}}. The statistics of the datasets are presented in \Cref{table:dataset}. 

\begin{table}[h] 
   \centering
   \caption{Statistics of datasets. \# Num stands for the number of numerical columns, and \# Cat stands for the number of categorical columns.} 
   \label{table:dataset}
   \resizebox{\columnwidth}{!}{
	\begin{tabular}{lcccccccc}
         \toprule[0.8pt]
         Dataset & \# Rows  & \# Num & \# Cat & \# Train & \# Validation & \# Test & Task  \\
         \midrule 
         Adult & $48,842$ & $6$ & $9$ & $28,943$ & $3,618$ & $16,281$ & Classification  \\
         Default & $30,000$ & $14$ & $11$ & $24,000$ & $3,000$ & $3,000$ & Classification   \\
         Shoppers & $12,330$ & $10$ & $8$ & $9,864$ & $1,233$ & $1,233$ & Classification   \\
         Magic & $19,019$ & $10$ & $1$ & $15,215$ & $1,902$ & $1,902$ & Classification  \\
         Beijing & $43,824$ & $7$ & $5$ & $35,058$ & $4,383$ & $4,383$ &  Regression   \\
         News & $39,644$ & $46$ & $2$ & $31,714$ & $3,965$ & $3,965$ & Regression \\
		\bottomrule[1.0pt] 
		\end{tabular}
   }
\end{table}

\section{Metrics}
\subsection{Shape}
Shape is proposed by SDMetrics~\cite{sdmetrics}\footnote{\url{https://docs.sdv.dev/sdmetrics}}. Shape calculates the column-wise density. Kolmogorov-Smirnov Test (KST)is used for numerical data and the Total Variation Distance (TVD) is used for categorical data. 

\paragraph{KST.} Given two continuous distributions $p_r(x)$ and $p_s(x)$, KST quantifies the distance between the two distributions using the upper bound of the discrepancy between two corresponding Cumulative Distribution Functions (CDFs):

\begin{align}
   \mathrm{KST}=\sup_{x}|F_r(x)-F_s(x)|
\end{align}

where $F_r(x)$ and $F_s(x)$ are the CDFs of $p_r(x)$ and $p_s(x)$, respectively:

\begin{align}
   F(x)=\int_{-\infty}^xp(x)dx
\end{align}

\paragraph{TVD.} TVD computes the frequency of each category value and expresses it as a probability. Then, the TVD score is the average difference between the probabilities of the categories:

\begin{align}
   \mathrm{TVD}=\dfrac{1}{2}\sum_{\omega\in\Omega}|R(\omega)-S(\omega)|
\end{align}

where $\omega$ represents all possible categories in a column $\Omega$. $R(\cdot)$ and $S(\cdot)$ denotes the real and synthetic
frequencies of these categories.

\subsection{Trend}
Trend is also proposed by SDMetrics. It calculates pair-wise column correlation. Pearson correlation is used for numerical data and contingency similarity is used for categorical data.

\paragraph{Pearson Correlation.} The Pearson correlation measures whether two continuous distributions are linearly correlated and is computed as

\begin{align}
   \rho_{x, y}=\dfrac{\mathrm{Cov}(x, y)}{\rho_x\rho_y}
\end{align}

where $x, y$ are two continuous columns, $\mathrm{Cov}$ is the covariance, and $\rho$ is the standard deviation. The performance of correlation estimation is measured by the average differences between the real data’s correlations and the synthetic data’s corrections

\begin{align}
   \mathrm{Pearson}=\dfrac{1}{2}\mathbb{E}_{x, y}|\rho^R(x, y)-\rho^S(x, y)|
\end{align}

where $\rho^R(x, y), \rho^S(x, y)$ represent the Pearson correlation between column $x$ and $y$ of the data $S$ or $R$, respectively.

\paragraph{Contingency similarity.} For a pair of categorical columns $A$ and $B$, the contingency similarity score computes the difference between the contingency tables using the TVD. The process is summarized by the formula below

\begin{align}
   \mathrm{Contingency}=\dfrac{1}{2}\sum_{\alpha\in A}\sum_{\beta\in B}|R_{\alpha,\beta}-S_{\alpha, \beta}|
\end{align}

where $\alpha, \beta$ represent all the possible categories in column $A, B$, respectively. $R_{\alpha,\beta}, S_{\alpha, \beta}$ are the joint frequency of $\alpha$ and $\beta$ in the data $R$ and $S$, respectively.

\section{Visualizations of Synthetic Data}
We present visualizations of the distributions of synthetic data generated by RecTable alongside real data.

\begin{figure}[!htbp]
\begin{minipage}[b]{0.48\linewidth}
\centering
\includegraphics[width=\linewidth]{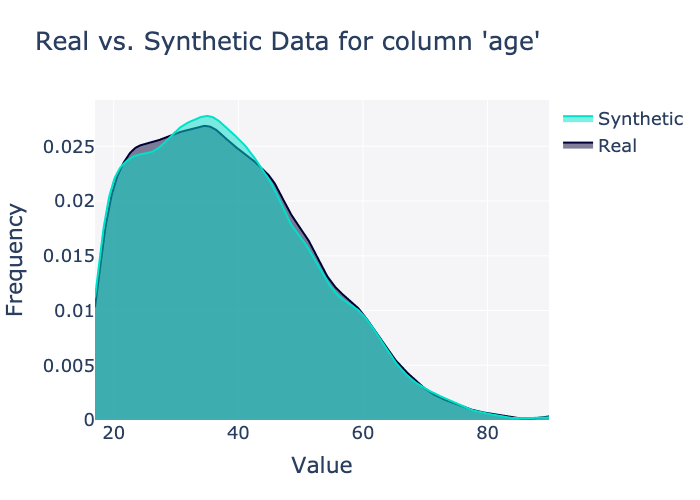}
\end{minipage}
\begin{minipage}[b]{0.48\linewidth}
\centering
\includegraphics[width=\linewidth]{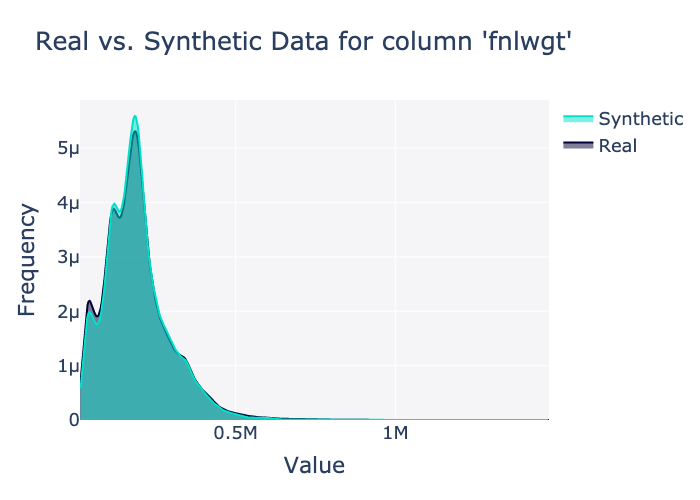}
\end{minipage}
\begin{minipage}[b]{0.48\linewidth}
\centering
\includegraphics[width=\linewidth]{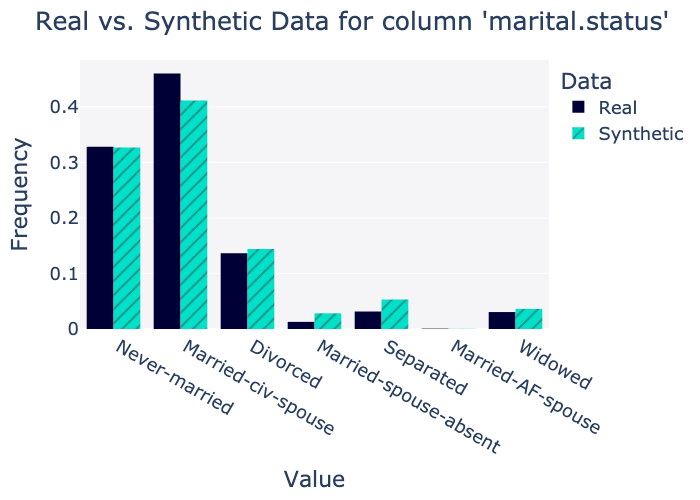}
\end{minipage}
\begin{minipage}[b]{0.48\linewidth}
\centering
\includegraphics[width=\linewidth]{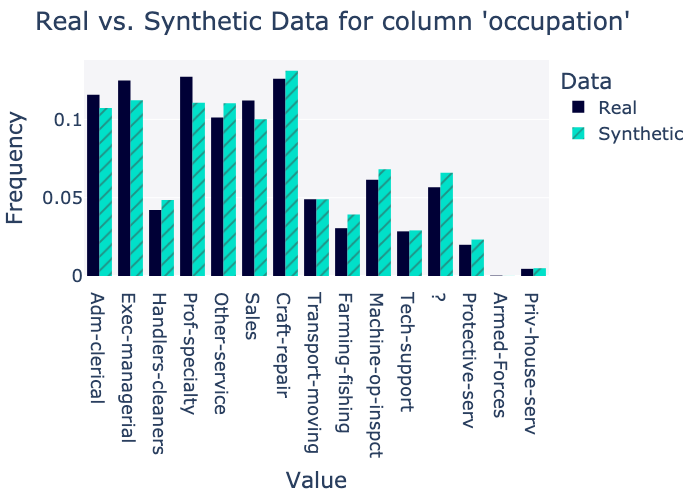}
\end{minipage}  
\caption{Visualizations of the generated and real adult dataset.}
\end{figure}

\begin{figure}[!htbp]
\begin{minipage}[b]{0.32\linewidth}
\centering
\includegraphics[width=\linewidth]{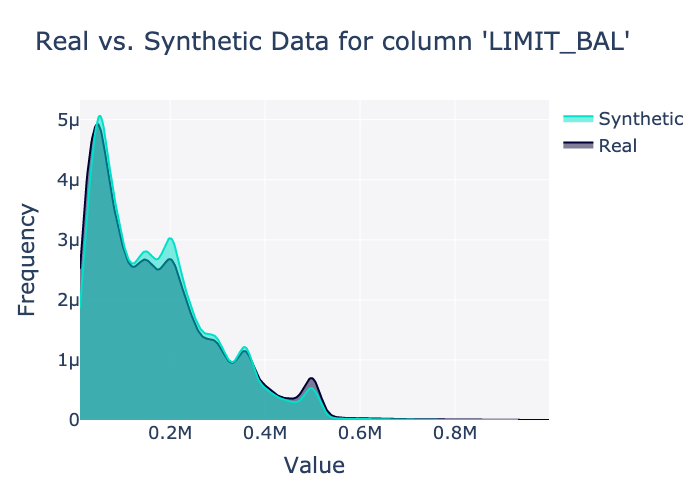}
\end{minipage}
\begin{minipage}[b]{0.32\linewidth}
\centering
\includegraphics[width=\linewidth]{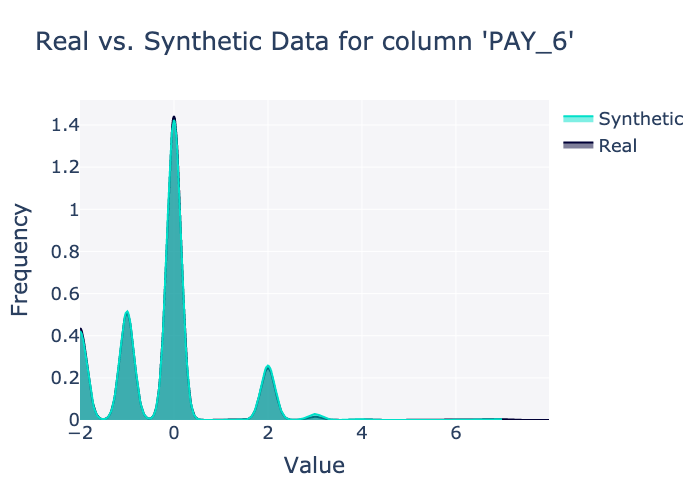}
\end{minipage}
\begin{minipage}[b]{0.32\linewidth}
\centering
\includegraphics[width=\linewidth]{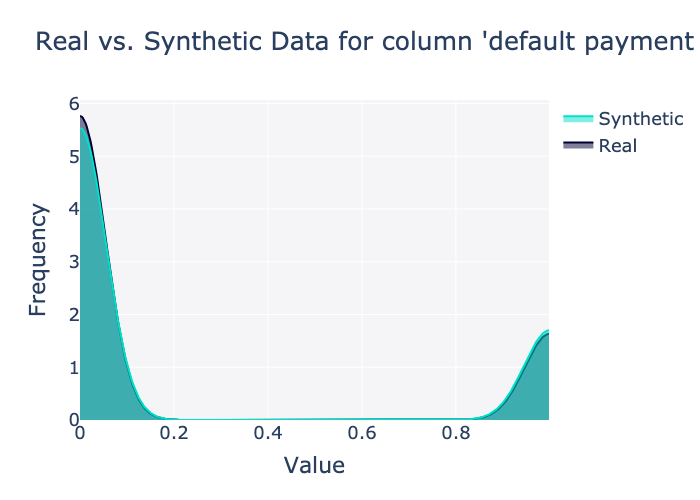}
\end{minipage} 
\caption{Visualizations of the generated and real default dataset.}
\end{figure}

\begin{figure}[!htbp]
\begin{minipage}[b]{0.48\linewidth}
\centering
\includegraphics[width=\linewidth]{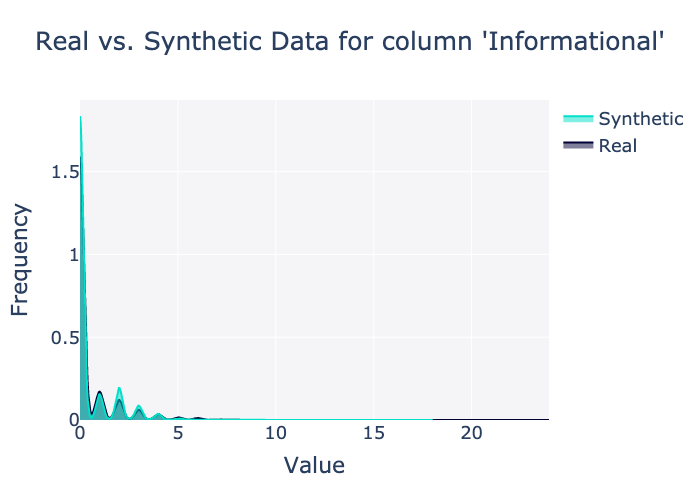}
\end{minipage}
\begin{minipage}[b]{0.48\linewidth}
\centering
\includegraphics[width=\linewidth]{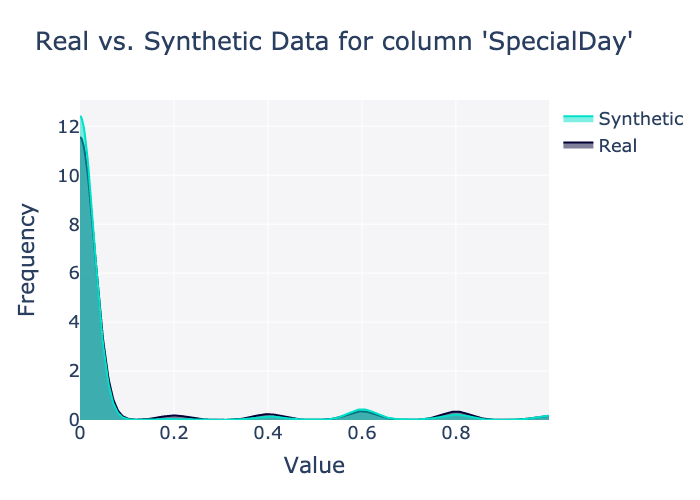}
\end{minipage}
\begin{minipage}[b]{0.48\linewidth}
\centering
\includegraphics[width=\linewidth]{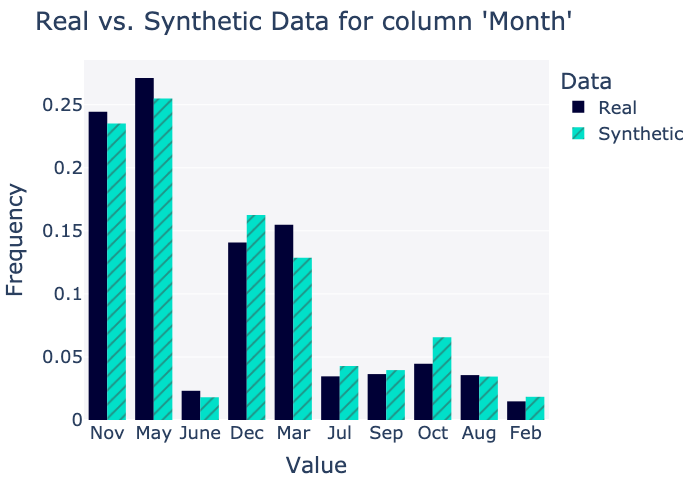}
\end{minipage}
\begin{minipage}[b]{0.48\linewidth}
\centering
\includegraphics[width=\linewidth]{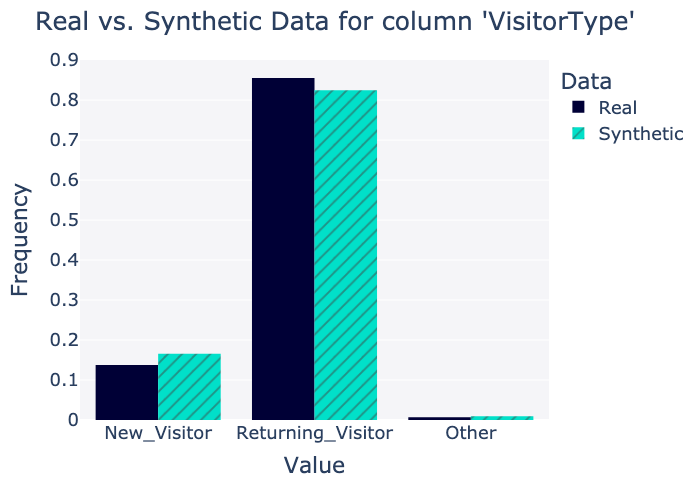}
\end{minipage}  
\caption{Visualizations of the generated and real shoppers dataset.}
\end{figure}

\begin{figure}[!htbp]
\begin{minipage}[b]{0.32\linewidth}
\centering
\includegraphics[width=\linewidth]{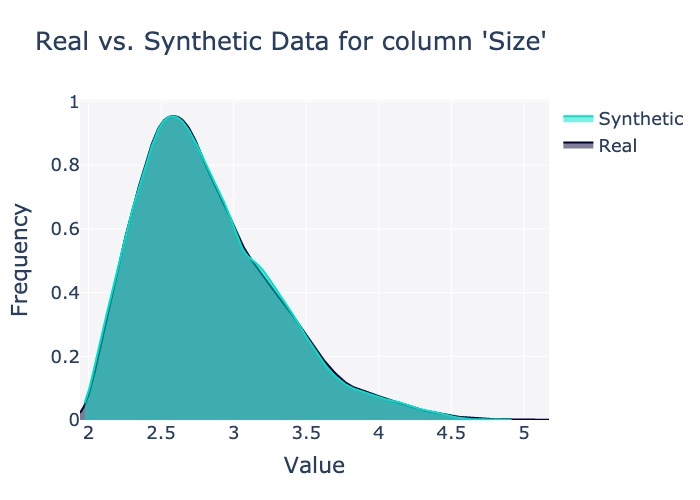}
\end{minipage}
\begin{minipage}[b]{0.32\linewidth}
\centering
\includegraphics[width=\linewidth]{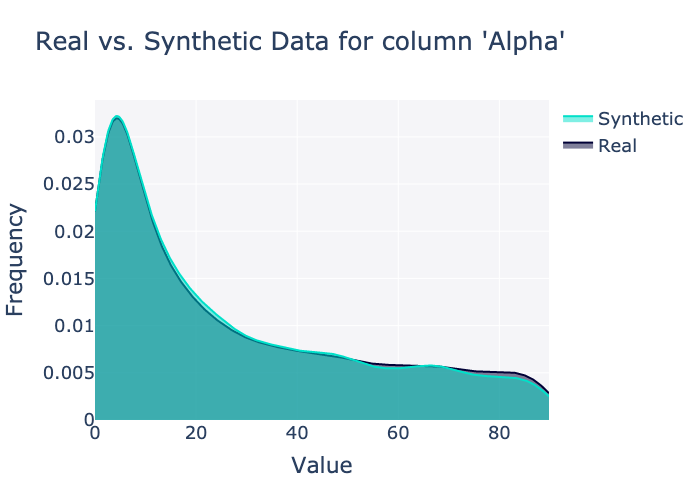}
\end{minipage}
\begin{minipage}[b]{0.32\linewidth}
\centering
\includegraphics[width=\linewidth]{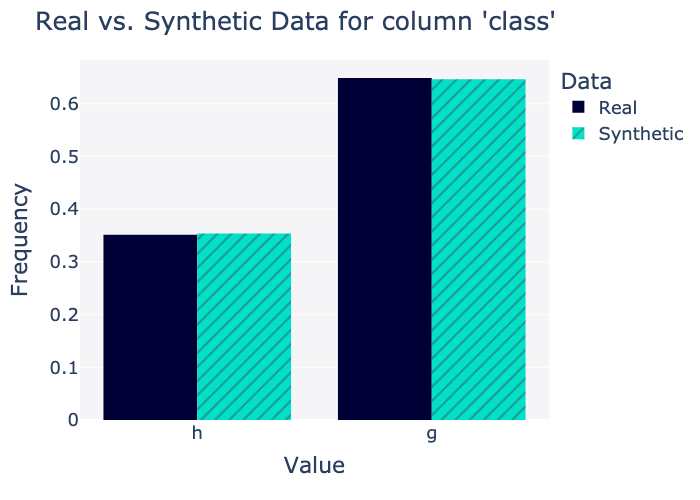}
\end{minipage}
\caption{Visualizations of the generated and real magic dataset.}
\end{figure}

\begin{figure}[!htbp]
\begin{minipage}[b]{0.32\linewidth}
\centering
\includegraphics[width=\linewidth]{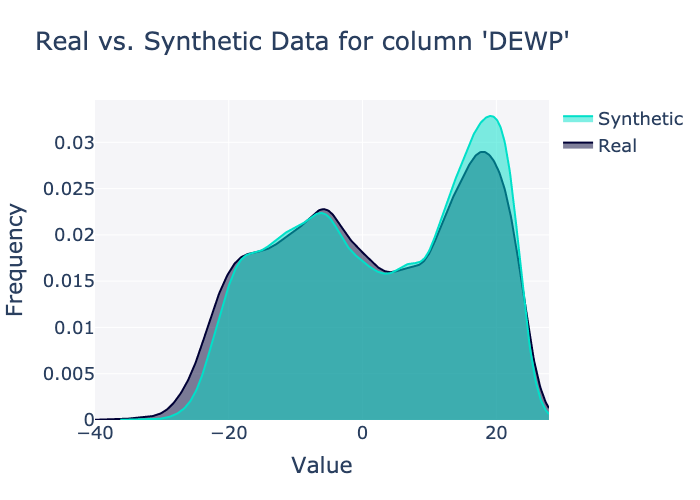}
\end{minipage}
\begin{minipage}[b]{0.32\linewidth}
\centering
\includegraphics[width=\linewidth]{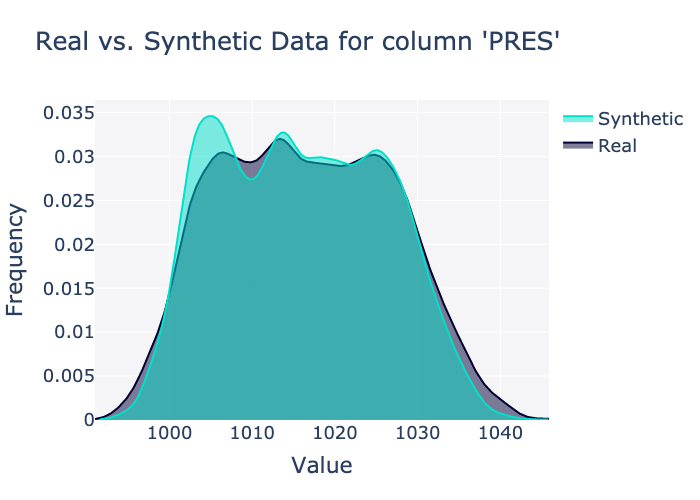}
\end{minipage}
\begin{minipage}[b]{0.32\linewidth}
\centering
\includegraphics[width=\linewidth]{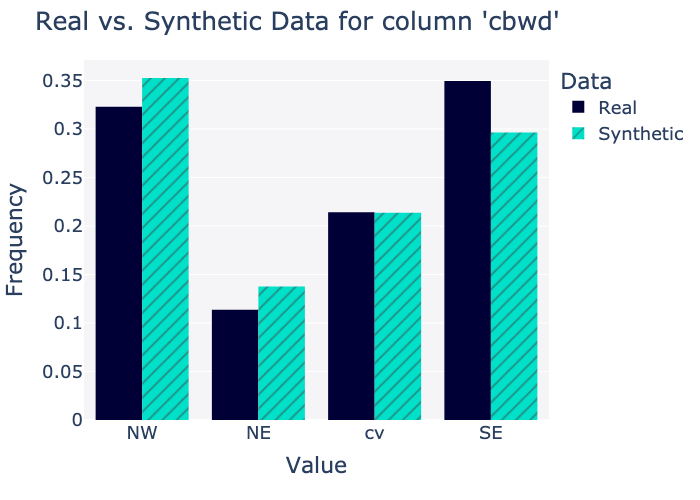}
\end{minipage}
\caption{Visualizations of the generated and real beijing dataset.}
\end{figure}

\begin{figure}[!htbp]
\begin{minipage}[b]{0.32\linewidth}
\centering
\includegraphics[width=\linewidth]{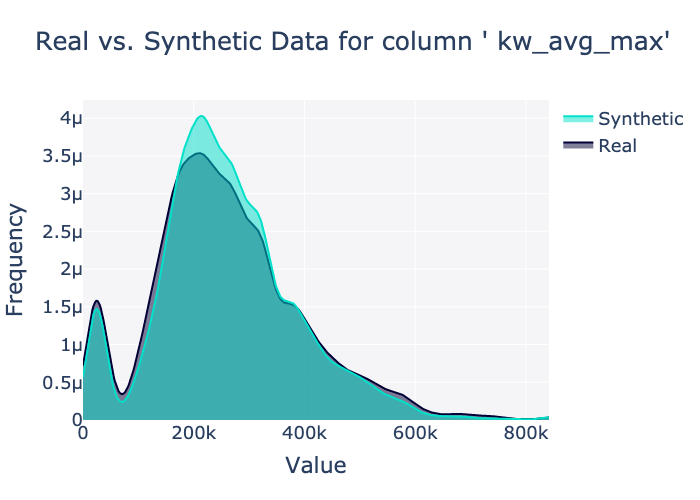}
\end{minipage}
\begin{minipage}[b]{0.32\linewidth}
\centering
\includegraphics[width=\linewidth]{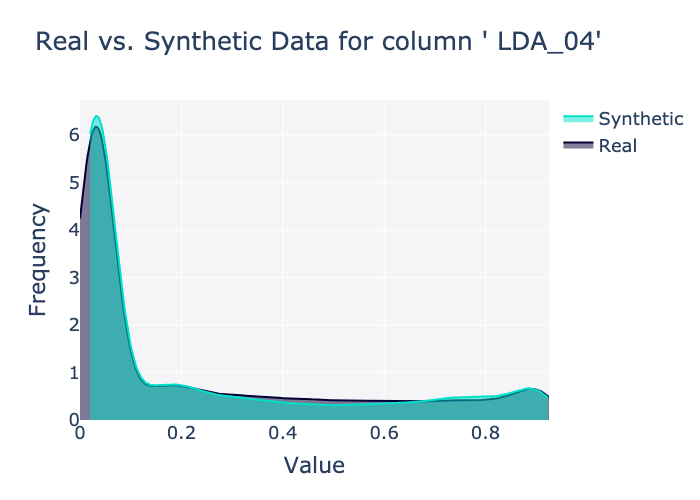}
\end{minipage}
\begin{minipage}[b]{0.32\linewidth}
\centering
\includegraphics[width=\linewidth]{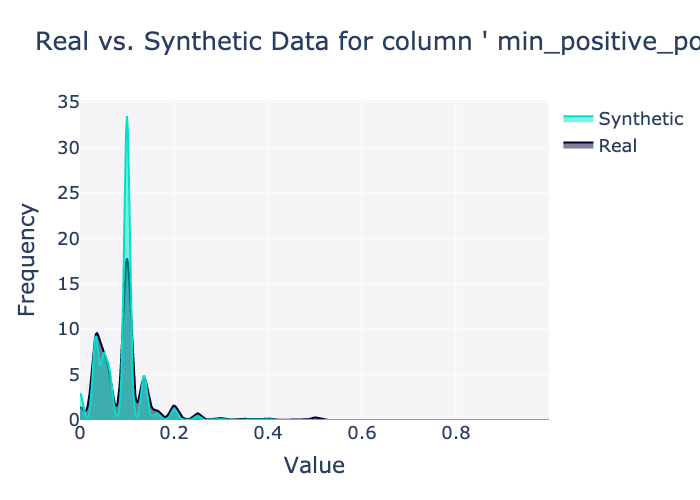}
\end{minipage}
\caption{Visualizations of the generated and real news dataset.}
\end{figure}

\end{document}